\documentclass{article} 
\usepackage{iclr2027_conference,times}

\usepackage{amsmath,amsfonts,bm}

\def\eqref#1{equation~\ref{#1}}

\def\1{\bm{1}}

\def\eps{{\epsilon}}

\DeclareMathAlphabet{\mathsfit}{\encodingdefault}{\sfdefault}{m}{sl}
\SetMathAlphabet{\mathsfit}{bold}{\encodingdefault}{\sfdefault}{bx}{n}

\newcommand{\E}{\mathbb{E}}

\newcommand{\R}{\mathbb{R}}

\usepackage{hyperref}
\hypersetup{colorlinks=true,linkcolor=blue!50!black,citecolor=blue!50!black,urlcolor=blue!50!black}
\usepackage{url}

\usepackage{amsmath,amssymb,amsthm,mathtools}
\usepackage{microtype}
\usepackage{xcolor}
\usepackage{booktabs}
\usepackage{algorithm,algpseudocode}
\usepackage{subcaption}
\usepackage{graphicx}
\usepackage{titletoc}
\usepackage{enumitem}
\usepackage{multirow}

\newtheorem{assumption}{Assumption}
\newtheorem{theorem}{Theorem}
\newtheorem{lemma}{Lemma}

\newtheorem{proposition}{Proposition}
\newtheorem{definition}{Definition}[]

\newcommand{\cB}{c_{\beta}}
\newcommand{\cF}{\mathcal{F}}
\newcommand{\Ocal}{\mathcal{O}}

\newcommand{\norm}[1]{\left\lVert#1\right\rVert}

\title{Why Does Adaptive Batching Help LLM \\ Pretraining? A Perspective from Unbounded \\Variance}

\author{Arda Fazla, Antesh Upadhyay, Ege C. Kaya, M. Berk Sahin \& Abolfazl Hashemi \\
Elmore Family School of Electrical and Computer Engineering\\
Purdue University\\
West Lafayette, IN 47907, USA \\
\texttt{\{afazla,aantesh,kayae,sahinm,abolfazl\}@purdue.edu}
}

\iclrfinalcopy 
\begin{document}

\maketitle
\lhead{Preprint}

\begin{abstract}
Increasing the batch size during training is a common practice in large language model (LLM) pretraining, yet the theoretical justification behind its success is not well understood. Analyses of stochastic optimization often assume uniformly bounded stochastic gradient variance, yet recent evidence suggests that this assumption fails in many practical nonconvex problems. The Blum--Gladyshev (BG-$0$) noise model relaxes this assumption by allowing the variance to grow quadratically with the distance from initialization, suggesting that batch size schedulers can help by controlling the variance growth during training. However, this growth can be overly conservative in practice. We empirically investigate variance growth in LLM pretraining and observe that a generalized BG model with a tunable growth exponent provides a tighter description of practical noise behavior. Motivated by this observation, we introduce the generalized BG-$a$ noise model, which interpolates between bounded variance ($a=0$) and BG-$0$ noise ($a=2$). Under $L$-smoothness, we derive an information-theoretic lower bound with growth-dependent oracle complexity $\Omega(\epsilon^{-(4+a)})$ and establish a matching upper bound in $\epsilon$-dependence by increasing the batch size as the iterates move away from initialization. Finally, we propose an adaptive batch scheduler that controls variance growth through dynamic batch size adjustments during training. In pretraining OLMo2 models of up to 1B parameters on C4, our scheduler achieves a lower validation loss than both small and large batch training under matched token budgets, while using less than 10\% of the iterations of small batch training.
\end{abstract}

\section{Introduction}
With the growing scale of training datasets, particularly in language model pretraining, large batch training is increasingly important for exploiting data parallelism. However, using a very large batch from initialization can reduce token efficiency, motivating schedules that begin with smaller batches and increase them as training progresses \citep{merrill2026critical}. Such schedules are already used in large scale training, e.g., GPT-3 gradually increased its batch size during the initial phase of pretraining \citep{brown2020language}. Existing adaptive batch schedulers commonly guide batch selection through gradient noise scale (GNS) estimates \citep{mccandlish2018empirical,naganuma2026adaptive}. However, \citet{merrill2026critical} show that this connection relies on assumptions about the optimizer and curvature that need not hold in deep learning, and find that GNS can be an unreliable proxy in practice. They instead estimate the critical batch size directly through additional branched training runs at different training stages. Despite the success of these schedules in practice, the theoretical justification for why increasing the batch size during training helps LLM pretraining is not well understood. In this work, we study this question from the perspective of stochastic gradient noise, where the variance grows as the iterates move away from initialization.

Classical optimization theory does not directly account for this behavior, as the convergence analysis of stochastic gradient methods commonly combines $L$-smoothness of the objective with a uniformly bounded variance assumption on the stochastic gradient oracle~\citep{moulines2011non,shamir2013stochastic,ghadimi2013stochastic}. For an unbiased estimator $g(x,\xi)$ of $\nabla f(x)$, this assumption requires $\mathbb{E}_{\xi}\|g(x,\xi)-\nabla f(x)\|^2\leq\sigma^2$ for every $x$, where $\sigma$ is constant and independent of the iteration. However, this assumption is often unrealistic because smoothness directly controls the changes in the population gradient, not the variance of its stochastic estimates. As a result, the bounded variance assumption need not hold even for unconstrained least squares problems \citep{alacaoglu2025towards, fazla2026lower, upadhyay2026beyond}, where differences between sample gradients and the population gradient can grow linearly with the parameters, causing variance to grow quadratically. As a remedy, researchers have analyzed the Blum--Gladyshev (BG-$0$) noise model \citep{blum1954approximation, gladyshev1965stochastic} which accommodates this behavior by allowing $\mathbb{E}_{\xi}\|g(x,\xi)-\nabla f(x)\|^2 \leq B^2\|x-x_0\|^2+G^2$, where $x_0$ is a fixed reference point, and $B,G$ are problem dependent constants. For smooth nonconvex optimization under this unbounded variance oracle, \citet{fazla2026lower} establish an $\Omega(\epsilon^{-6})$ stochastic first-order oracle lower bound for finding an $\epsilon$-stationary point and provide matching upper bounds for stochastic gradient descent (SGD) with dynamic batching. Their results show that distance-dependent noise incurs an intrinsic complexity penalty relative to the classical bounded variance rate of $\Theta(\epsilon^{-4})$ \citep{arjevani2023lower}.

\begin{figure}[!t]
    \centering
    \includegraphics[width=0.9\textwidth]{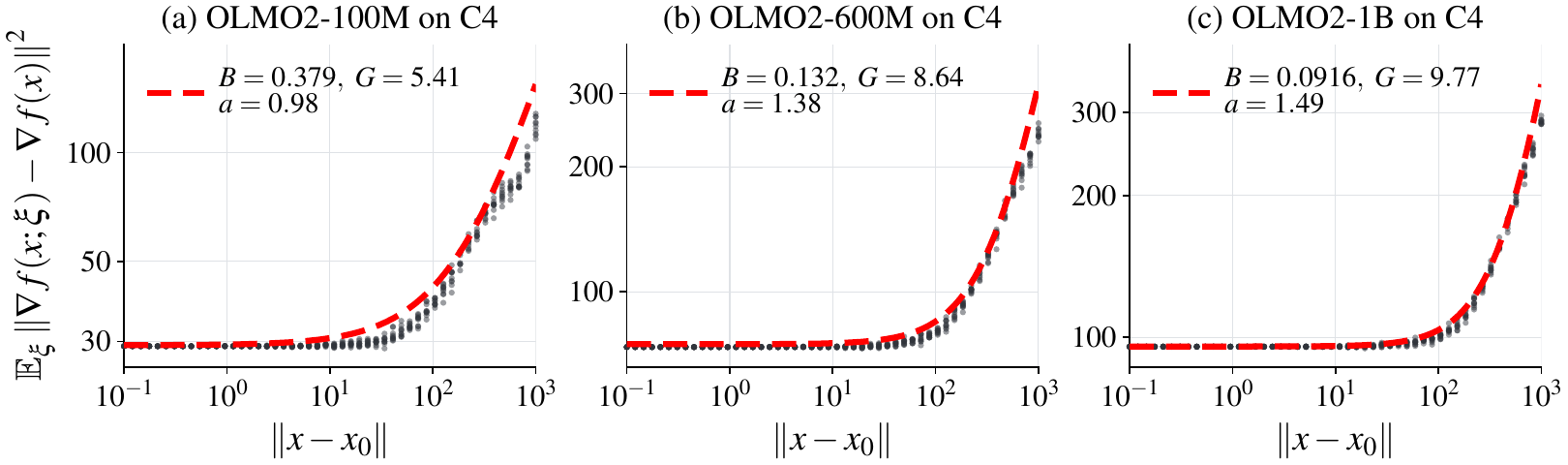}
    \caption{
    Empirical stochastic-gradient noise of OLMo2 models on C4 as a function of displacement from initialization. For each radius $r$, we evaluate $x=x_0+ru$ along ten independently sampled, normalized Gaussian directions and plot the corresponding minibatch variance estimates. The dashed red curve in each panel is the fitted BG-$a$ model, ($B^2\norm{x-x_0}^a+G^2$). The fitted parameters are: \textbf{(a)} OLMo2-100M: $B=0.379, G=5.41, a=0.98$; \textbf{(b)} OLMo2-600M: $B=0.132, G=8.64, a=1.38$; and \textbf{(c)} OLMo2-1B: $B=0.0916, G=9.77, a=1.49$. Across all three model sizes, the fitted curve captures both the initial noise floor governed by ($G^2$) and the distance-dependent growth governed by $B^2\norm{x-x_0}^a$, with the fitted exponent $a$ increasing with model size. The corresponding results for the vision models are provided in Appendix~\ref{app:vis_noise}.
    }
    \label{fig:olmo2_row}
\end{figure}

Although BG-$0$ relaxes the bounded variance assumption, its quadratic dependence on distance can overestimate the variance growth encountered in practice. As illustrated in Figure~\ref{fig:olmo2_row}, we find that the empirical variance growth in LLM pretraining is subquadratic, with fitted exponents $a\in[0.98,1.49]$ that increase with model size, and we observe a similar behavior for the vision models we study. Motivated by these observations, we introduce the generalized BG-$a$ noise model, $\mathbb{E}_{\xi}\|g(x,\xi)-\nabla f(x)\|^2 \leq B^2\|x-x_0\|^a+G^2$, which interpolates between uniformly bounded variance at $a=0$ and classical quadratic BG noise ($a=2$, denoted as BG-$0$ in prior works~\citep{fazla2026lower, upadhyay2026beyond}). For $L$-smooth nonconvex objectives, we establish matching lower and upper bounds with $\Theta(\epsilon^{-(4+a)})$ dependence for finding an $\epsilon$-stationary point for $a\in[0,2]$. Thus, the rate at which stochastic gradient noise grows with distance directly governs the oracle complexity, while recovering the bounded variance and BG-$0$ rates as the two endpoints.

One direct way to control distance-dependent noise is to increase the batch size as the iterates move far from initialization. Accordingly, \citet{fazla2026lower} use dynamic batching to control the minibatch variance under BG-$0$ noise, which we extend to the BG-$a$ model, where the corresponding batch size requirement scales with $B^2\|x_t-x_0\|^a+G^2$. We show that this dynamic batching strategy under the BG-$a$ model yields a matching $\Ocal(\epsilon^{-(4+a)})$ oracle complexity for SGD. Thus, we provide a theoretical explanation for increasing batch size during training: small batches are sufficient near initialization, where noise is small, while larger batches are needed to control variance growth as iterates move away from initialization. However, directly scaling batch sizes with noise level can lead to large per-iteration batch requirements as displacement grows. To alleviate this issue and make the method more flexible and practical, we extend the dynamically batched SGD analysis to SGD with momentum (SGDM) under BG-$a$ noise. Our analysis shows that momentum provides an additional form of averaging over time, reducing the sufficient per-iteration batch size at the cost of additional iterations while preserving the optimal $\Ocal(\epsilon^{-(4+a)})$ oracle complexity. This establishes a theoretical trade-off between batch size and iteration count under distance-dependent noise.

Together, these results offer a theoretical perspective on why increasing the batch size during pretraining works, complementary to empirical approaches based on GNS and critical batch size estimation. In particular, our BG-$a$ analysis connects the growth of the batch size to the growth of stochastic gradient noise, prescribing batch sizes that control variance growth as the iterates move away from initialization. To examine this perspective in practice, we translate the resulting variance control condition into an adaptive batch scheduler. In our LLM pretraining experiments, our scheduler achieves a lower validation loss than both small and large batch training under matched token budgets. We also provide experiments on image classification tasks, and discuss the role of adaptive batching in offline, multi-epoch training in Appendix~\ref{app:vis_exps}.

\textbf{We summarize our contributions as follows:}
\begin{itemize}[leftmargin=*,nosep]
    \item We empirically study the stochastic gradient noise in LLM pretraining and demonstrate that its variance grows subquadratically with the distance from initialization. We formalize this phenomenon by introducing the BG-$a$ oracle, which interpolates between bounded variance and BG-$0$ noise.
    \item For $L$-smooth nonconvex optimization, we establish an $\Omega(\epsilon^{-(4+a)})$ lower bound on the stochastic first-order oracle complexity for $a\in[0,2]$, characterizing how distance-dependent noise growth degrades the sample complexity of finding an $\epsilon$-stationary point.
    \item We establish matching $\Ocal(\epsilon^{-(4+a)})$ oracle complexity bounds for both SGD and SGDM under dynamic batching schedules. We identify momentum as a mechanism to reduce the required batch size at the expense of additional iterations, all while maintaining the optimal dependence on $\epsilon$.
    \item We instantiate our theoretical variance control analysis into a practical adaptive batch scheduler that scales batch sizes according to displacement from initialization and can be readily used in practice. In OLMo2 pretraining on C4, our scheduler achieves a lower validation loss than both small and large batch training under matched token budgets, while using less than $10\%$ of the iterations of small batch training.
\end{itemize}

\section{Gradient Noise Grows with Distance in LLM Pretraining}\label{sec:variance_ablation}

Before developing our theoretical results, we first examine the behavior of stochastic gradient noise in LLM pretraining. Specifically, we study how the gradient variance grows as the model parameters move away from initialization and assess how tightly the quadratic distance dependence assumed by the BG-$0$ oracle captures this growth. We then investigate whether allowing a tunable growth exponent yields a tighter characterization of the observed variance.

We study language modeling with OLMo2-100M, OLMo2-600M, and OLMo2-1B models \citep{olmo20242olmo2furious} on C4 \citep{c4}. To study the effect of displacement on noise growth without following an optimizer generated trajectory, we evaluate the model at randomly perturbed parameter states, without performing training. Starting from a randomly initialized anchor point $x_0$, we construct evaluation points as
$$
x=x_0+ru,\qquad
u=\frac{Dz}{\norm{Dz}},\qquad
z\sim\mathcal N(0,I),
$$
where $r>0$ is the distance from initialization and $D$ is a diagonal matrix specifying the relative perturbation scales. We set the diagonal entries of $D$ to $2$ for parameters of the LM head and $1$ for all remaining parameters. Thus, we double the Gaussian perturbation scale of the LM head before globally normalizing the direction, preserving $\norm{x-x_0}=r$. For each model, we consider $50$ logarithmically spaced radii and sample $10$ independent directions at each radius.

This modeling choice is motivated by our observation that the LM head exhibits the largest weight changes among all parameter groups in our pretraining runs. This observation agrees with prior work that identifies a gradient bottleneck at the LM head \citep{godey2026lost} and highlights heterogeneous output layer gradients and, hence, larger effective learning rates for the LM head weights \citep{glentis2026revisiting}. We use a factor of two as a simple way to reflect this behavior, so that the perturbation directions more closely resemble how the parameters move during training.

At each evaluation point $x$, we estimate the variance of stochastic gradients around a fixed large batch reference gradient $\nabla\tilde f(x)$, which is computed before the evaluation process and used as an approximation to the population gradient. Specifically, we approximate the stochastic gradient variance $\E_\xi\norm{\nabla f(x;\xi)-\nabla f(x)}^2$ using
$\frac1M\sum_{i=1}^M
\norm{g_b^{(i)}(x)-\nabla\tilde f(x)}^2, \text{where} \ g_b^{(i)}(x)=\frac1b\sum_{j=1}^b\nabla f(x;\xi_{i,j}).$
For all OLMo2 models, we use a reference batch of $512$ sequences and $M=32$ stochastic batches, each containing $4$ sequences, where sequence length is set as $2048$.

\begin{figure}[!t]
    \centering
    \includegraphics[width=0.8\textwidth]{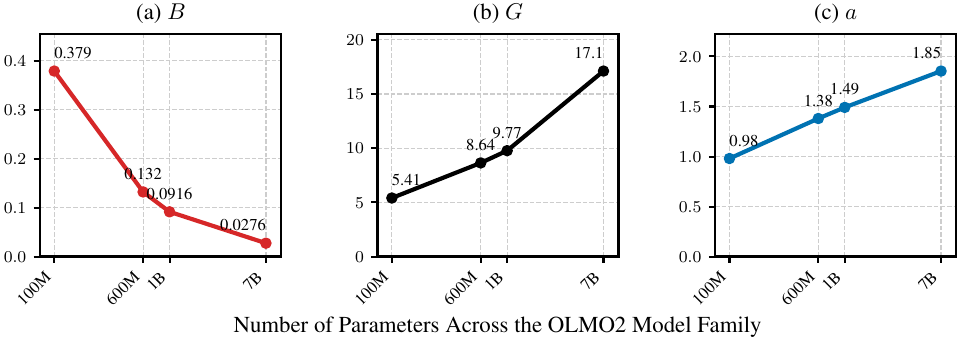}
    \caption{
    Estimated BG-$a$ noise-model parameters across OLMo2 model sizes. We observe that as model size increases, (a) $B$ tends to decrease, while (b) $G$ and (c) $a$ tend to increase.
    }
    \label{fig:bga_ablation}
\end{figure}

Figure~\ref{fig:olmo2_row} shows the distribution of individual variance measurements across the three OLMo2 models, together with the best fitting curves of the form $B^2\norm{x-x_0}^a+G^2$, subject to $B,G\geq 0$ and $a\in[0,2]$. The fitted exponents are approximately $a=0.98$, $1.38$, and $1.49$ for OLMo2-100M, OLMo2-600M, and OLMo2-1B, respectively. These results indicate that the distance from initialization is a useful predictor of gradient noise growth. Across the sampled points, the proposed BG-$a$ condition closely captures the observed increase in variance as the iterates move farther from initialization. Moreover, allowing the exponent $a$ to adapt to the observed noise-growth behavior provides a better fit than the BG-$0$ condition of \cite{alacaoglu2025towards, fazla2026lower, upadhyay2026beyond}, which fixes the corresponding exponent to $2$. Figure~\ref{fig:bga_ablation} further shows how the fitted parameters change with model size: as the model grows, the growth exponent $a$ and the noise floor $G$ increase, while $B$ decreases. We observe a similar subquadratic noise growth for ResNet50 on image classification tasks, which we report in Appendix~\ref{app:vis_noise}.

These observations motivate the BG-$a$ noise model, which we formalize and analyze in the next section, and an adaptive batch size scheduler that increases the batch size as the iterates move farther from initialization, thereby controlling variance growth throughout training.

\section{Theoretical Analysis under the BG-\texorpdfstring{$a$}{a} Oracle}\label{sec:theory_main}

\subsection{Preliminaries}
For a given random variable $X$, we write $\E[X]$ to denote the expectation with respect to all randomness introduced by the algorithm. $\|\cdot\|$ denotes the Euclidean norm, and $\langle \cdot,\cdot\rangle$ denotes the Euclidean inner product. The notation $\Ocal(\cdot)$ suppresses numerical constants that are independent of the target accuracy $\eps$. We use $\cF_t$ to denote the history of the algorithm up to iteration $t$, which contains the initialization and all samples used at iterations $0,\ldots,t-1$. We denote the corresponding conditional expectation by $\E_t[\cdot]=\E[\cdot\mid\cF_t].$ In assumptions involving an oracle sample, $\E_\xi$ denotes expectation only
with respect to the sample $\xi$, with the query points held fixed.

We consider the stochastic optimization problem $\min_{x \in \mathbb{R}^d} f(x), \ f(x):=\mathbb{E}_{\xi}\bigl[f(x;\xi)\bigr].$ Here, $f:\mathbb{R}^d \to \mathbb{R}$ is differentiable, nonconvex, and bounded below by $f^{\inf}:=\inf_x f(x)>-\infty$. We denote the initialization query point with $x_0$ and use $\Delta=f(x_0)-f^{\inf} \ge 0$.

The convergence criterion that we consider is the average expected squared gradient norm along the iterates $x_0,\ldots,x_{T-1}$,
\begin{equation}\label{eq:main_stationarity}
    \frac1T\sum_{t=0}^{T-1}\E\norm{\nabla f(x_t)}^2 \le \eps^2.
\end{equation}
Equivalently, returning $x_\tau$ with $\tau$ drawn uniformly from $\{0,\ldots,T-1\}$ yields $\mathbb{E}\|\nabla f(x_\tau)\|^2\le\epsilon^2$. Given a query point $x$, the stochastic first-order oracle (SFO) returns a stochastic gradient $\nabla f(x;\xi)$. We measure the computational complexity by the total number of SFO calls. At iteration $t$, the algorithms we consider query the oracle $b_t$ times at $x_t$ and form the minibatch gradient
\begin{equation}\label{eq:main_minibatch}
    g_t=\frac1{b_t}\sum_{i=1}^{b_t}\nabla f(x_t;\xi_{t,i}),
\end{equation}
where the batch size $b_t$ is chosen based on $\cF_t$ before sampling, and the samples $\xi_{t,1},\ldots,\xi_{t,b_t}$ are drawn independently of each other and of $\cF_t$. Hence, iteration $t$ uses $b_t$ SFO calls. We next state the assumptions used in our work.

\subsection{Assumptions}
\begin{assumption}[$L$-smoothness of $f$]\label{as:l_smooth}
The objective function $f$ is globally $L$-smooth with $L>0$ such that:
$$
\|\nabla f(x)-\nabla f(y)\|\le L\|x-y\|
\qquad\text{for all }x,y\in\R^d.
$$
\end{assumption}

\begin{assumption}[Unbiased BG-$a$ stochastic oracle]\label{as:bg_a_oracle}
Given $a\ge0$, for every $x \in \mathbb{R}^d$, the stochastic gradient oracle is unbiased, and there exist constants $B,G \ge 0$ satisfying:
$$
\mathbb{E}_{\xi}\bigl[\nabla f(x;\xi)\bigr] = \nabla f(x),
\qquad
\mathbb{E}_{\xi}
\left\|\nabla f(x;\xi)-\nabla f(x)\right\|^2
\le B^2\norm{x-x_0}^a+G^2.
$$
\end{assumption}

\subsection{Convergence Results}
We now summarize our three main theoretical results. Complete statements, constants, and proofs are given in the Appendix.

\begin{theorem}[Lower bound under BG-$a$ noise]\label{thm:main_lower_bound}
Fix any finite $a\ge0$. Under Assumption~\ref{as:l_smooth}, for all $\eps>0$ satisfying $\eps^2\le c\,L\Delta$, where $c>0$ is an absolute constant specified in Appendix~\ref{app:lower-bound}, every algorithm that returns a point $\widehat x$ satisfying $\E\norm{\nabla f(\widehat x)}^2\le\eps^2$ for every unbiased BG-$a$ oracle must incur worst-case expected SFO complexity
\begin{equation}\label{eq:main_lower_rate}
    \Omega_a\!\left(
    \frac{L\Delta}{\eps^2}
    +\frac{G^2L\Delta}{\eps^4}
    +\frac{B^2L\Delta^{a+1}}{\eps^{4+a}}
    \right).
\end{equation}
\end{theorem}

For Theorem~\ref{thm:main_lower_bound}, the oracle may be any unbiased vector estimator satisfying the BG-$a$ variance bound. See Appendix \ref{app:lower-bound} for the complete statement and proof.

\begin{theorem}[SGD Upper Bound with Dynamic Batching]\label{thm:main_sgd}
Suppose Assumptions~\ref{as:l_smooth} and~\ref{as:bg_a_oracle} hold with $0\le a\le2$. Fix $\sigma^2>0$ and let $0<\eps^2\le\min\{2\sigma^2,L\Delta\}$. At iteration $t$, use
\begin{equation}\label{eq:main_sgd_parameters}
    b_t=\max\!\left\{1,
    \left\lceil\frac{B^2\norm{x_t-x_0}^a+G^2}{\sigma^2}\right\rceil\right\},
    \qquad
    \eta=\frac{\eps^2}{2L\sigma^2},
    \qquad
    T=\left\lceil\frac{4\Delta}{\eta\eps^2}\right\rceil,
\end{equation}
and update the iterates using $x_{t+1}=x_t-\eta g_t$, where $g_t$ is the minibatch gradient in \eqref{eq:main_minibatch}. Then \eqref{eq:main_stationarity} holds with the expected SFO cost given as
\begin{equation}\label{eq:main_sgd_cost}
    \E\!\left[\sum_{t=0}^{T-1}b_t\right]
    =\Ocal\!\left(
    L\Delta(\sigma^2+G^2)\eps^{-4}
    +B^2L\Delta^{a+1}\eps^{-(4+a)}
    \right).
\end{equation}
In particular, for fixed $\sigma^2$, the expected SFO complexity is $\Ocal(\eps^{-(4+a)})$. The complete theorem statement and proof are provided in Appendix~\ref{sec:convergence_of_sgd}.
\end{theorem}

\begin{theorem}[SGDM Upper Bound with Dynamic Batching]\label{thm:main_sgdm}
Suppose Assumptions~\ref{as:l_smooth} and~\ref{as:bg_a_oracle} hold with $0\le a\le2$. Fix $\beta\in[0,1)$ and $\sigma^2>0$, and let $\cB=\frac{1-\beta}{1+\beta}$ and $0<\eps^2\le\min\{8\sigma^2(1-\beta^2),L\Delta\}$. At iteration $t$, use
\begin{equation}\label{eq:main_sgdm_parameters}
\begin{aligned}
    b_t&=\max\!\left\{1,
    \left\lceil\frac{\cB(B^2\norm{x_t-x_0}^a+G^2)}{\sigma^2}\right\rceil\right\},\\
    \eta&=\frac{\cB\eps^2}{16L\sigma^2},
    \qquad
    T=\left\lceil\frac{16\Delta}{\eta\eps^2}\right\rceil,
\end{aligned}
\end{equation}
and, with $g_t$ as in \eqref{eq:main_minibatch}, update the iterates using
\begin{equation}\label{eq:main_sgdm_update}
    m_{-1}=0,
    \qquad
    m_t=\beta m_{t-1}+(1-\beta)g_t,
    \qquad
    x_{t+1}=x_t-\eta m_t.
\end{equation}
Then \eqref{eq:main_stationarity} holds, and
\begin{equation}\label{eq:main_sgdm_cost}
    \E\!\left[\sum_{t=0}^{T-1}b_t\right]
    =\Ocal\!\left(
    L\Delta\left(\frac{\sigma^2}{\cB}+G^2\right)\eps^{-4}
    +B^2L\Delta^{a+1}\eps^{-(4+a)}
    \right).
\end{equation}
Thus, for fixed $\beta$ and $\sigma^2$, SGDM also uses $\Ocal(\eps^{-(4+a)})$ expected SFO calls. The complete theorem statement and proof are provided in Appendix~\ref{sec:convergence_of_sgdm}.
\end{theorem}

When $B>0$, the upper bounds match the growth-dominated lower-bound exponent: $\eps^{-4}$ at $a=0$, $\eps^{-(4+a)}$ for $0<a<2$, and $\eps^{-6}$ at $a=2$. When $B=0$ and $G>0$, both stochastic rates reduce to $\eps^{-4}$. We emphasize that the restriction $a\le2$ in the upper bounds is needed to control the $a$-th displacement moment using the second moment supplied by the variance assumption. We discuss the obstruction that occurs when $a>2$ in Appendix~\ref{sec:convergence_of_sgd}.

\subsection{Implications for Batch Size Scheduling}\label{sec:theory_implications}
Theorems~\ref{thm:main_sgd} and~\ref{thm:main_sgdm} prescribe a batch size that tracks the BG-$a$ variance bound at the current iterate. For $a>0$, since $x_0$ is the initialization, the prescribed batch size starts at $b_0=\max\{1,\lceil\cB G^2/\sigma^2\rceil\}$, with $\cB=1$ for SGD, and increases only as the iterates move away from initialization, in proportion to $B^2\norm{x_t-x_0}^a$. When $a=0$, with the convention $\|x-x_0\|^0=1$, the variance is uniformly bounded and the prescribed batch size remains constant. In our proofs, the expected squared displacement $\E\norm{x_t-x_0}^2$ is bounded by a quantity that increases with $t$, so the prescribed batch size is small at the start of training and can grow as training progresses. This avoids the drawbacks of the two fixed batch alternatives: a fixed small batch does not control the variance growth as the iterates move away from initialization, which requires a smaller step size and more iterations, whereas a fixed large batch chosen for the variance level late in training spends unnecessary samples early in training, where the noise is small. Moreover, the batch size rule depends on the noise only through the parameters $(B,G,a)$, which we estimate empirically in Section~\ref{sec:variance_ablation}, and serves as the basis of our adaptive batch scheduler in Section~\ref{sec:methodology}. Finally, momentum reduces the prescribed batch size by a factor of $\cB$ at the same query point, at the cost of a $1/\cB$ times larger iteration count, providing a trade-off between batch size and iteration count while preserving the $\Ocal(\eps^{-(4+a)})$ oracle complexity.

\section{Adaptive Batch Size Scheduler}\label{sec:methodology}

We introduce an adaptive batch size scheduler that dynamically adjusts the batch size throughout training to account for the BG-$a$ noise growth. Motivated by the empirical evidence in Section~\ref{sec:variance_ablation} and the theoretical analysis in Section~\ref{sec:theory_main}, our scheduler determines the batch size at each iteration based on the displacement of the current iterate from the initialization $x_0$. We first describe the batch size rule and then present the optimization scheme built on top of it.

Under the BG-$a$ noise model, the gradient variance at iteration $t$ is bounded by $B^2\norm{x_t-x_0}^a+G^2$, where the exponent $a$ governs how the variance grows with the distance from initialization, while $B$ and $G$ control the distance-dependent and constant noise components, respectively. Our scheduler takes these noise model parameters as inputs, which we estimate using the noise measurements in Section~\ref{sec:variance_ablation}, and determines the batch size before sampling the stochastic gradients at each iteration. Following Theorem~\ref{thm:main_sgdm}, given a target noise scale $\sigma^2>0$ and a momentum parameter $\beta\in[0,1)$ with $\cB=(1-\beta)/(1+\beta)$, we define the distance-dependent function
\begin{equation}\label{eq:method_batch_score}
v_t := \frac{\cB\left(B^2\norm{x_t-x_0}^a+G^2\right)}{\sigma^2}.
\end{equation}
For simplicity, throughout this section we assume $B,G>0$ and $a \in (0,2]$. Hence, $v_0=\cB G^2/\sigma^2>0$ and $v_t \geq v_0$.

We initialize the batch size as $b_0 \geq v_0$ and restrict subsequent batch sizes to power-of-two multiples of $b_0$ to reduce the frequency of potentially costly batch size adjustments, while allowing distributed systems to naturally benefit from data parallelism. At each iteration, we round the ratio $v_t/v_0$ upward to the smallest power of two that is at least this ratio:
\begin{equation}\label{eq:method_batch_rule}
k_t:=2^{\left\lceil\log_2(v_t/v_0)\right\rceil}, \qquad b_t:=k_tb_0.
\end{equation}
This guarantees $k_t\in\{1,2,4,8,\ldots\}$, $k_0=1$, and $v_t/v_0\le k_t<2v_t/v_0$. Thus, $b_t\ge v_t$, and conditional independence and the BG-$a$ bound give
\begin{equation}\label{eq:method_relative_variance}
\mathbb{E}_t\norm{g_t-\nabla f(x_t)}^2
\le
\frac{B^2\norm{x_t-x_0}^a+G^2}{b_t}
= \frac{\sigma^2v_t}{\cB k_tb_0}
\le \frac{\sigma^2v_0}{\cB b_0}
=\frac{G^2}{b_0}
\le\frac{\sigma^2}{\cB}.
\end{equation}
The scheduler therefore keeps the minibatch variance bound at or below its initial level. Since $b_t$ is a positive integer and $b_t\ge v_t$, it satisfies the batch size requirement in Theorem~\ref{thm:main_sgdm}, and in Theorem~\ref{thm:main_sgd} when $\beta=0$ so that $\cB=1$.

While $\cB$ and $\sigma^2$ determine the initial batch size $b_0$, they cancel out in the ratio $v_t/v_0$. Hence, the multiplier $k_t$ can be computed directly from $(x_t,x_0,B,G,a)$ as
\begin{equation}\label{eq:adaptive_batch_size_function}
k_t=\mathrm{AdaptiveBatchSize}(x_t,x_0,B,G,a):=
2^{\left\lceil\log_2\!\left(1+\dfrac{B^2}{G^2}\norm{x_t-x_0}^a\right)\right\rceil}.
\end{equation}
Note that $k_t$ depends on the noise parameters only through the ratio $B^2/G^2$ and the exponent $a$. In particular, rescaling $B^2$ and $G^2$ by a common factor, such as the minibatch size used when estimating the variance in Section~\ref{sec:variance_ablation}, leaves $k_t$ unchanged.

Algorithm~\ref{alg:adaptive_batching_optimizer} applies the proposed batch size scheduler to an optimizer of choice, with a base learning rate schedule $\{\eta_t\}_{t=0}^{T-1}$ specified for the initial batch size $b_0$. We adjust the learning rate to the selected batch size using the optimizer-dependent scaling rules of linear scaling for SGD-based optimizers \citep{goyal2017accurate,mccandlish2018empirical} and square-root scaling for Adam \citep{malladi2022sdes,merrill2026critical}. Using the multiplier $k_t=b_t/b_0$, we set
\begin{equation}\label{eq:batch_lr_scaling}
\bar\eta_t=\eta_tf_{\mathrm{OPT}}(k_t),
\qquad
f_{\mathrm{OPT}}(k)=
\begin{cases}
k, & \mathrm{OPT}\in\{\mathrm{SGD},\mathrm{SGDM}\},\\
\sqrt{k}, & \mathrm{OPT} \in \{\mathrm{Adam}\}.
\end{cases}
\end{equation}
Consequently, doubling the batch size doubles the learning rate for SGD and SGDM and multiplies it by $\sqrt{2}$ for Adam.

\begin{algorithm}[H]
\caption{Optimization with the Adaptive Batch Size Scheduler}\label{alg:adaptive_batching_optimizer}
\begin{algorithmic}[1]
\Require Initialization $x_0$; initial batch size $b_0$; optimizer $\mathrm{OPT}$ with its hyperparameters; base learning rate schedule $\{\eta_t\}_{t=0}^{T-1}$; iteration budget $T\ge1$; noise parameters $B,G>0,a\in(0,2]$
\State $s_0\gets\mathrm{OPT.Initialize}()$ \Comment{Initialize optimizer state}
\For{$t=0,1,\ldots,T-1$}
    \State $k_t\gets\Call{AdaptiveBatchSize}{x_t,x_0,B,G,a}$ \Comment{Using \eqref{eq:adaptive_batch_size_function}}
    \State $b_t\gets k_tb_0$
    \State $\bar\eta_t\gets\eta_tf_{\mathrm{OPT}}(k_t)$ \Comment{Using \eqref{eq:batch_lr_scaling}}
    \State Draw fresh conditionally independent samples $\xi_{t,1},\ldots,\xi_{t,b_t}$
    \State $g_t\gets\dfrac1{b_t}\sum_{i=1}^{b_t}\nabla f(x_t;\xi_{t,i})$
    \State $(x_{t+1},s_{t+1})\gets\mathrm{OPT.Step}(x_t,s_t,g_t,\bar\eta_t)$
\EndFor
\State \Return $x_T$
\end{algorithmic}
\end{algorithm}

In Algorithm~\ref{alg:adaptive_batching_optimizer}, $s_t$ corresponds to the optimizer state at iteration $t$, such as momentum for SGDM, or first and second moment estimates for Adam, whereas SGD requires no states. The scheduler changes the batch size and adjusts the learning rate, while the optimizer follows its usual state update.

Our analysis requires $b_0\geq v_0=\cB G^2/\sigma^2$. However, in practice, a suitable target noise level $\sigma^2$ is not known a priori and can differ across model-dataset pairs. Therefore, in our experiments, we set $b_0$ equal to the batch size of the small batch baseline, so that BG-$a$ and the small batch baseline start from the same batch size and the same base learning rate. We choose this as the smallest batch size that trains stably and attains a competitive final loss after tuning the learning rate through a grid search (see Appendix~\ref{app:exp_details}).

The scheduler follows the batch size rule based on our theoretical analysis. Since $b_t\ge v_t$ at every iteration, the resulting batch sizes satisfy the variance control condition in Theorems~\ref{thm:main_sgd} and~\ref{thm:main_sgdm}, while the power-of-two rounding enables better compatibility with common pretraining frameworks, e.g., OLMo-core \citep{olmo20242olmo2furious}. As discussed in Section~\ref{sec:theory_implications}, the scheduler starts training with a small batch, where the noise is small, and increases the batch size as the iterates move away from initialization and the variance grows. The noise parameters $(B,G,a)$ used by the scheduler are estimated through the measurements in Section~\ref{sec:variance_ablation}, connecting our empirical observations and theoretical analysis to a practical scheduler. We discuss further practical considerations in Appendix~\ref{sec:discussion}. Next, we present experiments showing that, under matched token budgets, our scheduler achieves a lower validation loss in LLM pretraining while using far fewer iterations than small batch training.

\section{Experiments}\label{sec:experiments}
We evaluate our adaptive batch size scheduler in LLM pretraining, where large batch training is particularly important for exploiting data parallelism, and where each training token is seen at most once (online training). Our experiments on image classification, where models are trained for multiple epochs (offline training), are provided in Appendix~\ref{app:vis_exps}. We pretrain OLMo2-100M, OLMo2-600M, and OLMo2-1B models on C4 using Chinchilla optimal training budgets of approximately $20$ tokens per parameter \citep{hoffmann2022training}, corresponding to $2$B, $12$B, and $20$B tokens, respectively. All runs use AdamW with a linear warmup over the first $5\%$ of training followed by a cosine decay. We compare our scheduler, denoted as BG-$a$, against a small and a large fixed batch baseline under the same token budget. BG-$a$ starts from the small batch size, uses the noise parameters $(B,G,a)$ estimated in Section~\ref{sec:variance_ablation}, and scales the learning rate with the square-root rule in \eqref{eq:batch_lr_scaling}. Further training details are provided in Appendix~\ref{app:exp_details}.

\begin{figure}[!t]
    \centering
    \includegraphics[width=0.9\textwidth]{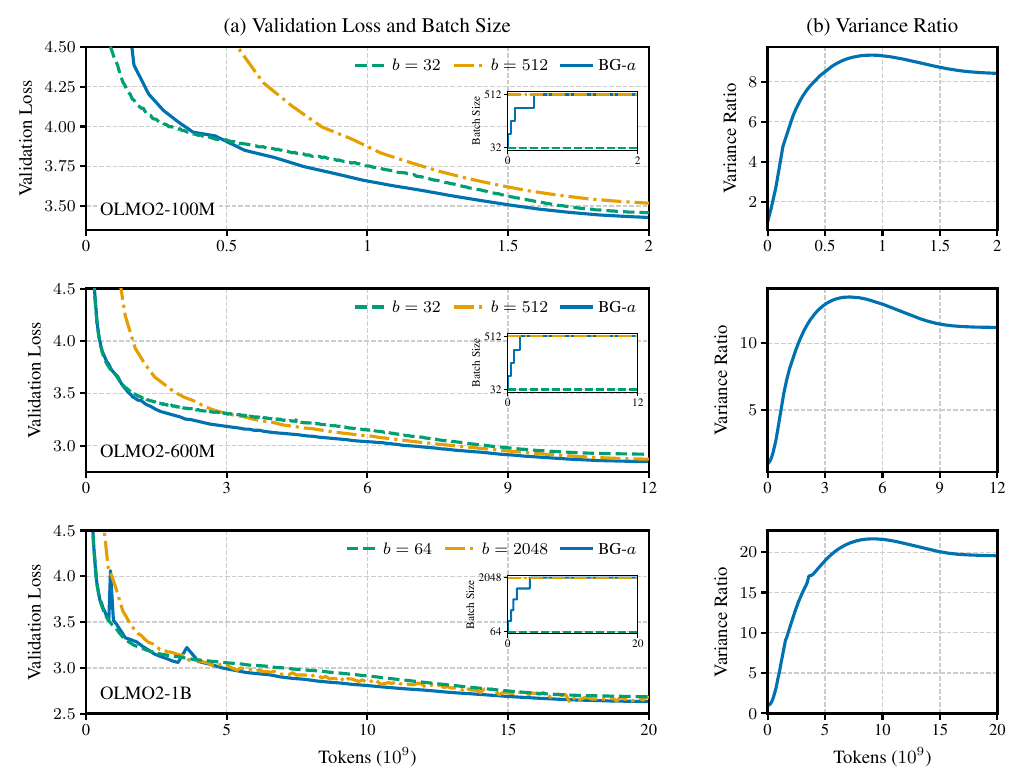}
    \caption{
        Results for OLMo2-100M (top), OLMo2-600M (middle) and OLMo2-1B (bottom) pretrained on C4.
        (a) Validation loss for a fixed small batch ($b=32$ for 100M and 600M, $b=64$ for 1B), a fixed large batch ($b=512$ for 100M and 600M, $b=2048$ for 1B), and BG-$a$, with batch size across iterations shown in the inset.
        (b) Modeled variance ratio for BG-$a$, with parameters $B=0.379, G=5.41, a=0.98$ (100M), $B=0.132, G=8.64, a=1.38$ (600M) and $B=0.0916, G=9.77, a=1.49$ (1B).
    }
    \label{fig:olmo2}
\end{figure}

Table~\ref{tab:olmo2_batching} and Figure~\ref{fig:olmo2} show that BG-$a$ achieves the lowest validation loss across all $3$ model sizes. Hence, we observe that the optimization benefit of adaptive batching directly translates into a lower validation loss. We emphasize that BG-$a$ achieves these results while using only $8.76\%$, $8.10\%$, and $5.25\%$ of the iterations of small batch training for OLMo2-100M, OLMo2-600M, and OLMo2-1B, respectively. Since the scheduler starts from the small batch size and increases it only as the iterates move away from initialization, it uses more iterations than large batch training, but it remains within the same order of magnitude. The reduction in iterations also translates into a shorter training time, with BG-$a$ reducing the wall-clock time of small batch training by approximately $40\%$ for OLMo2-100M, $28\%$ for OLMo2-600M and $20\%$ for OLMo2-1B (Table~\ref{tab:repro_lm}).

The comparison between the two fixed batch baselines further supports our analysis on the BG-$a$ noise model. For OLMo2-100M, whose fitted exponent is $a=0.98$, small batch training outperforms large batch training. For OLMo2-600M and OLMo2-1B, where the fitted exponents are larger, small batch training attains the worst validation loss among the three methods. This is consistent with our theory, as a faster growth of the variance with the distance from initialization makes a fixed small batch less effective in controlling the noise later in training. Note that as detailed in Appendix~\ref{app:exp_details}, we select the base learning rate through a grid search for small batch training, and obtain the learning rates for large batch and adaptive batch training through square-root scaling, without additional tuning. This setup favors small batch training, as it is the only configuration whose learning rate is tuned directly. Similarly, the small batch size itself is selected as the smallest batch size that trains stably and attains a competitive final loss in each setting, as also described in Appendix~\ref{app:exp_details}. We adopt this protocol to ensure that the improvements of our scheduler do not result from an insufficient learning rate or insufficiently small batch size. We also present downstream accuracy scores on HellaSwag and CommonsenseQA in Table~\ref{tab:olmo2_downstream} in Appendix \ref{app:add_exps}. We provide a detailed discussion in Appendix \ref{sec:discussion}.

\section{Conclusion}\label{sec:conclusion}
In this work, we studied why increasing the batch size during training is effective in LLM pretraining, a common practice that lacks a clear theoretical justification. We approached this question from the perspective of stochastic gradient noise whose variance grows as the iterates move away from initialization, a behavior that the classical bounded variance assumption does not capture. Motivated by our empirical observations in language model pretraining, we introduced the BG-$a$ noise model, which interpolates between bounded variance and BG-$0$ noise, and characterized how the growth of the noise governs the oracle complexity of finding a stationary point. We showed that increasing the batch size as the iterates move away from initialization offers a theoretical explanation for batch size schedules. We argued that small batches are sufficient early in training, while larger batches are needed later to control the growing noise. We also theoretically demonstrated that momentum provides a trade-off between the batch size and the number of iterations. Building on this analysis, we proposed an adaptive batch size scheduler that improves upon both small and large batch training in LLM pretraining. We hope that this perspective encourages further work connecting stochastic optimization theory with the design of batch size schedules in large scale training.

\begin{table}[!t]
\centering
\caption{Validation performance and optimization cost of OLMo2
models pretrained on C4. Validation loss is reported as
mean$\pm$std across $3$ seeds, with iteration percentages relative
to small batch runs.}
\label{tab:olmo2_batching}
\small
\setlength{\tabcolsep}{4pt}
\begin{tabular}{llrrr}
\toprule
Model & Batch Size & Validation Loss & Iterations ($T$)
& Iterations (\%) \\
\midrule
\multirow{3}{*}{OLMo2-100M}
 & Small ($b=32$) & $3.4609 \pm 0.0037$ & 30,518   & 100.00\% \\
 & Large ($b=512$) & $3.5139 \pm 0.0071$ & 1,908    & 6.25\% \\
 & BG-$a$            & $3.4347 \pm 0.0085$ & 2,673  & 8.76\% \\
\midrule
\multirow{3}{*}{OLMo2-600M}
 & Small ($b=32$) & $ 2.9188 \pm 0.0062 $ & 183,106 & 100.00\% \\
 & Large ($b=512$) & $ 2.8710 \pm 0.0079 $ & 11,444 & 6.25\% \\
 & BG-$a$            & $ 2.8482 \pm 0.0098 $ & 14,824 & 8.10\% \\
\midrule
\multirow{3}{*}{OLMo2-1B}
 & Small ($b=64$) & $ 2.6845 \pm 0.0034 $ & 152,588 & 100.00\% \\
 & Large ($b=2048$) & $ 2.6696 \pm 0.0061 $ & 4,769 & 3.13\% \\
 & BG-$a$            & $ 2.6328 \pm 0.0095 $ & 8,015 & 5.25\% \\
\bottomrule
\end{tabular}
\end{table}

\newpage

\bibliography{iclr2027_conference}
\bibliographystyle{iclr2027_conference}

\clearpage
\appendix

\startcontents[appendix]
\printcontents[appendix]{}{1}{%
  \section*{\Large Appendix Table of Contents}
  \setcounter{tocdepth}{2} 
}

\newpage
\setcounter{theorem}{0}

\title{Why Does Adaptive Batching Help LLM Pretraining? A Perspective from Unbounded Variance}
\author{}
\date{}

\maketitle

\section{Related Work}\label{app:related_work}

\textbf{Large scale training.} Larger batches enable greater data parallelism, but their benefits greatly depend on the workload and the optimizer. Extensive experiments conducted by \cite{shallue2019measuring} show that the reduction in training steps from increasing batch size varies substantially across tasks and depends on hyperparameter tuning and compute budgets. Learning rate scaling is also hugely dependent on the optimizer and training scheme. \cite{goyal2017accurate, mccandlish2018empirical} use linear scaling for large batch SGD, whereas \cite{malladi2022sdes, merrill2026critical} apply square root scaling for AdamW and RMSprop. In large language model pretraining, the critical batch size (CBS) \citep{mccandlish2018empirical} marks the scale beyond which additional parallelism comes at the expense of token efficiency. Batch size warmup, where training starts with a small batch that is increased as training progresses, is used in large scale pretraining \citep{brown2020language} and can improve token efficiency compared to training with a large batch from initialization \citep{merrill2026critical}. \cite{zhang2025does} find that the CBS depends more strongly on training data size than model size. \cite{merrill2026critical} identify limitations of the gradient noise scale as a proxy for critical batch size and estimate it directly through branched training runs. These findings directly motivate our displacement dependent adaptive batch scheduler in language model pretraining, as we couple batch size changes with optimizer-dependent learning rate scaling rules and compare training quality under matched sample or token budgets. While these works approach adaptive batching from the perspective of training efficiency and predicting the critical batch size during training, our theoretical analysis of the BG-$a$ noise model and the batch size rule derived from it offer a complementary theoretical perspective on why increasing the batch size during training helps.

\textbf{Effect of batch size.} Batch size controls both the accuracy of stochastic gradients and the cost of each update, which has motivated researchers to come up with methods to adjust the batch sizes throughout training. Early approaches increase batch size to maintain a suitable gradient signal-to-noise ratio \citep{bollapragada2018adaptive}. As batch size and learning rate jointly influence optimization, the relation between them has been studied by \cite{balles2016coupling} and \cite{smith2017don}, where it is shown that increasing the batch size can have similar effects to learning rate decay in common training schedules. Theoretical work also studies the computational and communication benefits of increasing batches in nonconvex SGD \citep{yu2019computation}, and the effect of batch size adaptation based on past stochastic gradients in variance reduced methods \citep{ji2020history}. From the empirical side of things, \cite{mccandlish2018empirical} propose the gradient noise scale (GNS) as a useful proxy to the critical batch size during training. \cite{naganuma2026adaptive} extend the GNS framework to cover non Euclidean algorithms such as sign and spectral descent, and their variants.

\textbf{Online and offline learning.} The role of stochastic gradient noise differs between offline training, where the model is trained for multiple epochs over a fixed dataset, and online training, where each sample is seen at most once. In the offline setting, small batch training is known to generalize better than large batch training, which \cite{keskar2016large} attribute to large batches converging to sharper minima. This has led to the view of SGD noise as an implicit regularizer. \cite{vyas2023beyond} show that this implicit bias advantage disappears in the online setting, where the benefit of small batches becomes purely computational and lower noise runs consistently perform better once the number of gradient steps is controlled. They further hypothesize that online SGD takes noisy steps along the trajectory of noiseless gradient descent. LLM pretraining, which is typically performed for a single epoch over a large corpus, falls into this online regime. This distinction is consistent with the difference we observe between our language model and image classification experiments. Our analysis concerns the optimization of the training objective. In online LLM pretraining, each batch consists of previously unseen tokens, so the training and validation losses estimate the same population loss, and our scheduler attains a lower validation loss than small batch training. In multi-epoch image classification, our scheduler achieves a lower training loss than small batch training, whereas small batch training retains its generalization advantage in test loss.

\textbf{Stochastic optimization and variance models.} The classical analysis of smooth nonconvex optimization often assumes an unbiased stochastic gradient oracle with uniformly bounded variance. Under this setting, SGD achieves $\Ocal(\epsilon^{-4})$ oracle complexity for finding a point satisfying $\mathbb{E}\|\nabla f(x)\|^2\leq\epsilon^2$ \citep{ghadimi2013stochastic} matching the lower bound of \cite{arjevani2023lower}. Several analyses have attempted to relax the uniform variance bound, e.g., the expected smoothness framework of \cite{khaled2020better} controls the second moment of the stochastic gradient through objective suboptimality and the gradient norm. The heavy-tail framework \citep{zhang2020adaptive} analyzes gradient clipping under bounded moments of order below two, allowing the variance to be infinite. The BG-$0$ oracle allows variance to grow quadratically with distance from a fixed reference point, building on classical assumptions in stochastic approximation \citep{blum1954approximation, gladyshev1965stochastic}. \cite{alacaoglu2025towards} revisit this condition and develop convergence guarantees for convex optimization without requiring bounded domains. For smooth nonconvex objectives, \cite{fazla2026lower} establish an $\Omega(\epsilon^{-6})$ lower bound under BG-$0$ and obtain matching upper bounds through dynamic batching. \cite{upadhyay2026beyond} analyze BG-$0$ under both standard and symmetric generalized smoothness, achieving $\Ocal(\epsilon^{-6})$ oracle complexity with normalized SGD without increasing batch sizes. Existing analyses of the BG-$0$ oracle assume quadratic growth of the gradient variance with distance from a reference point, while the empirical validity of this assumption in practical training settings remains unexplored. Our work considers variance growth of order $\norm{x-x_0}^a$ for $a\in[0,2]$ and characterizes how the growth exponent determines the optimal dependence of oracle complexity on accuracy. We also assess our proposed BG-$a$ noise model empirically in language model pretraining, with additional vision experiments in Appendix~\ref{app:vis_exps}, and use the observed relationship between displacement and gradient variance to explain why increasing the batch size during training helps and to guide adaptive batch size selection.

\section{Lower bound under \texorpdfstring{$L$}{L}-Smoothness}\label{app:lower-bound}

We build on the lower bound in Theorem 1 of \cite{fazla2026lower} with the oracle model
\begin{equation}\label{eq:model}
\E_\xi[\nabla f(x;\xi)]=\nabla f(x),\qquad
\E_\xi\norm{\nabla f(x;\xi)-\nabla f(x)}^2
\le B^2\norm{x-x_0}^a+G^2,
\end{equation}
where we use the convention $r^0=1$ for all $r\ge0$, including $r=0$. Here $a=2$ gives BG-0, while $a=0$ gives bounded variance. We consider lower bounded, globally $L$-smooth objectives with $f(x_0)-f^{\inf}\le\Delta$, where $f^{\inf}:=\inf_x f(x)$. We use $\nabla f(x;\xi)$ for the stochastic oracle calls. For the proof of the lower bound, this notation denotes a general unbiased vector estimator and does not require a sample loss whose gradient equals the estimator. We write $g(x;\xi):=\nabla f(x;\xi)$ when referring to an objective--oracle pair $(f,g)$. Each call uses a fresh independent seed and counts as one SFO call. 


\subsection{Definitions and auxiliary results}\label{app:lower_auxiliary}

Before getting into the proof details, we restate the definitions and chain properties used in the proof. The deterministic chain and masked oracle are those of \cite{arjevani2023lower}, and the range bound below is adopted from \cite{fazla2026lower}, which is used to shift the chain to a nonpositive function.

\begin{definition}[Progress function]\label{def:lower_progress}
For $w\in\R^T$ and $\gamma\ge0$, define
$$
\operatorname{prog}_\gamma(w)
:=\max\bigl(\{i\in\{1,\ldots,T\}:|w_i|>\gamma\}\cup\{0\}\bigr).
$$
This is the largest coordinate index whose absolute value exceeds the threshold, and is defined to be zero if no such coordinate exists. E.g., $\operatorname{prog}_1(w)=T$ means that $|w_T|>1$, without imposing any condition on the preceding coordinates.
\end{definition}

\begin{definition}[Zero-respecting algorithm]\label{def:lower_zero_respecting}
Let $\operatorname{supp}(v):=\{i:v_i\ne0\}$. Suppose we initialize with $x_0=0$ and $g_t$ is the stochastic gradient returned at query $x_t$. An algorithm is zero-respecting if, almost surely, each query made before termination satisfies
$$
\operatorname{supp}(x_t)
\subseteq\bigcup_{j=0}^{t-1}\operatorname{supp}(g_j).
$$
For the lower bound in Stage 1 below, we also require its output after $\tau$ oracle calls to satisfy
$$
\operatorname{supp}(\widehat x)
\subseteq\bigcup_{j=0}^{\tau-1}\operatorname{supp}(g_j).
$$
These restrictions allow a coordinate to become nonzero only after it has appeared in an oracle response.
\end{definition}

\begin{definition}[Deterministic chain]\label{def:lower_chain}
For $T\ge1$ and $w\in\R^T$, define
\begin{equation}\label{eq:lower_chain_definition}
\bar f_T(w):=-\Psi(1)\Phi(w_1)
+\sum_{i=2}^T\bigl[\Psi(-w_{i-1})\Phi(-w_i)-\Psi(w_{i-1})\Phi(w_i)\bigr],
\end{equation}
where
$$
\Psi(r):=\begin{cases}
0,&r\le1/2,\\
\exp\!\left(1-\dfrac1{(2r-1)^2}\right),&r>1/2,
\end{cases}
\qquad
\Phi(r):=\sqrt{e}\int_{-\infty}^r e^{-z^2/2}\,dz.
$$
These scalar functions define the base chain used in the construction of the lower bound.
\end{definition}

\begin{lemma}[Deterministic chain properties (Lemma 2 of \cite{arjevani2023lower} restated)]\label{lem:lower_chain_properties}
Setting $\ell_1=152$ and $\varsigma=23$, the function in Definition~\ref{def:lower_chain} satisfies the following properties with:
\begin{enumerate}
\item $\bar f_T(0)-\inf_w\bar f_T(w)\le12T$.
\item Its gradient is globally $\ell_1$-Lipschitz and, for every $w\in\R^T$,
\begin{equation}\label{eq:chainbounds}
\operatorname{Lip}(\nabla\bar f_T)\le\ell_1,
\qquad \norm{\nabla\bar f_T(w)}_\infty\le\varsigma.
\end{equation}
\item The gradient can extend the chain by at most one coordinate:
$$
\operatorname{prog}_0(\nabla\bar f_T(w))
\le\operatorname{prog}_{1/2}(w)+1.
$$
\item If $\operatorname{prog}_1(w)<T$, then
$$
\left|\partial_{\operatorname{prog}_1(w)+1}\bar f_T(w)\right|>1,
\qquad \norm{\nabla\bar f_T(w)}_\infty>1.
$$
\end{enumerate}
\end{lemma}

\begin{lemma}[Range bound from ~\cite{fazla2026lower}]\label{lem:lower_chain_range}
Let $M_T:=\sup_w\bar f_T(w)$. Then $M_T$ is finite and for every $w\in\R^T$,
\begin{equation}\label{eq:range}
-24T\le\bar f_T(w)-M_T\le0.
\end{equation}
\end{lemma}
\begin{proof}
The definitions above give $0\le\Psi\le e$, $\Psi(1)=1$, and $0\le\Phi\le\sqrt{2\pi e}$. For every real $r$, at most one of $\Psi(r)$ and $\Psi(-r)$ is nonzero. Thus each summand with $i\ge2$ in \eqref{eq:lower_chain_definition} has absolute value at most $e\sqrt{2\pi e}$, and the first term has absolute value at most $\sqrt{2\pi e}$. Hence
$$
|\bar f_T(w)|\le\sqrt{2\pi e}+(T-1)e\sqrt{2\pi e}
\le Te\sqrt{2\pi e}<12T.
$$
It follows that $M_T\le12T$ and $\inf_w\bar f_T(w)\ge-12T$, proving \eqref{eq:range}.
\end{proof}

\begin{definition}[Bernoulli masked chain oracle]\label{def:lower_masked_oracle}
Fix $p\in(0,1]$ and draw $\xi\sim\mathrm{Bernoulli}(p)$. For $w\in\R^T$, we define $\bar g_T(w;\xi)\in\R^T$ (coordinate wise) by
\begin{equation}\label{eq:lower_masked_definition}
[\bar g_T(w;\xi)]_i
:=\partial_i\bar f_T(w)
\left[1+\mathbf1_{\{i>\operatorname{prog}_{1/4}(w)\}}(\xi/p-1)\right],
\qquad i=1,\ldots,T.
\end{equation}
Each oracle call uses a fresh independent seed. Coordinates beyond the progress index are suppressed when $\xi=0$ and multiplied by $1/p$ when $\xi=1$.
\end{definition}

\begin{lemma}[Masked oracle properties from Lemma 3 of \cite{arjevani2023lower}]\label{lem:lower_masked_properties}
For every $w\in\R^T$, the oracle in Definition~\ref{def:lower_masked_oracle} satisfies
\begin{equation}\label{eq:chainvariance}
\E_\xi[\bar g_T(w;\xi)]=\nabla\bar f_T(w),\qquad
\E_\xi\norm{\bar g_T(w;\xi)-\nabla\bar f_T(w)}^2
\le\varsigma^2\frac{1-p}{p}.
\end{equation}
Writing $k=\operatorname{prog}_{1/4}(w)$, for both possible seeds its support satisfies
$$
\operatorname{supp}(\bar g_T(w;\xi))
\subseteq\{1,\ldots,\min(T,k+1)\},\qquad
\operatorname{supp}(\bar g_T(w;0))\subseteq\{1,\ldots,k\}.
$$
Thus, a response can extend progress by at most one coordinate, and such an extension requires $\xi=1$. For each fixed seed, these support statements hold simultaneously for every $w$, yielding the probability-$p$ zero-chain property used in \cite{arjevani2023lower}.
\end{lemma}
\begin{proof}
Since $\E_\xi[\xi/p]=1$, we achieve unbiasedness by taking expectation in \eqref{eq:lower_masked_definition}. Furthermore, $\operatorname{prog}_{1/2}(w)\le k$, hence, Lemma~\ref{lem:lower_chain_properties} implies $\partial_i\bar f_T(w)=0$ for $i>k+1$. When $\xi=0$, the mask also removes coordinate $k+1$, which proves the support statements.

If $k<T$, the only possibly nonzero component of the oracle error is
$$
[\bar g_T(w;\xi)-\nabla\bar f_T(w)]_{k+1}
=\partial_{k+1}\bar f_T(w)(\xi/p-1).
$$
Using $\E_\xi[(\xi/p-1)^2]=(1-p)/p$ and applying the gradient bound in Lemma~\ref{lem:lower_chain_properties} gives
$$
\E_\xi\norm{\bar g_T(w;\xi)-\nabla\bar f_T(w)}^2
=|\partial_{k+1}\bar f_T(w)|^2\frac{1-p}{p}
\le\varsigma^2\frac{1-p}{p}.
$$
If $k=T$, the error is zero as the mask does not change any coordinates. Hence, the same bound holds.
\end{proof}

\subsection{Lower-bound theorem and proof}

We prove the lower bound under the weaker requirement $\E\norm{\nabla f(\widehat x)}\le\eps$. Since $\mathbb{E}\|\nabla f(\widehat x)\|^2\le\epsilon^2$ implies this requirement by Cauchy--Schwarz, the lower bound also applies to the squared gradient criterion in Theorem~\ref{thm:main_lower_bound}.

\begin{theorem}[Lower bound under smoothness]\label{thm:lower_bound}
Under fixed $L,\Delta>0$, $B,G\ge0$, and a finite $a\ge0$, let $\ell=155$ and $\varsigma=23$, and suppose
\begin{equation}\label{eq:small}
0<\eps^2\le\frac{L\Delta}{98304\ell}.
\end{equation}
Any algorithm that returns $\widehat x$ after a random number $\tau$ of SFO calls and satisfies
\begin{equation}\label{eq:criterion}
\E\norm{\nabla f(\widehat x)}\le\eps
\end{equation}
for every admissible instance must satisfy
\begin{equation}\label{eq:explicit}
\sup_{(f,g)}\E[\tau]
\ge\frac{L\Delta}{786432\ell\eps^2}
\left[1+\frac{G^2+B^2(\Delta/(16\eps))^a}
{64\varsigma^2\eps^2}\right].
\end{equation}
Consequently, the worst-case expected SFO complexity is
\begin{equation}\label{eq:rate}
\Omega_a\!\left(\frac{L\Delta}{\eps^2}
+\frac{G^2L\Delta}{\eps^4}
+\frac{B^2L\Delta^{a+1}}{\eps^{4+a}}\right).
\end{equation}
\end{theorem}

\begin{proof}
\textbf{Proof roadmap and relation to prior works.}
We first explain the roadmap that we follow for the proof and explain its relation to \cite{arjevani2023lower} and PASTA \citep{fazla2026lower}. We follow PASTA's travel coordinate and activation construction. This choice is intuitive, as PASTA's variance assumption (BG-$0$) is a special case of the BG-$a$ noise model with $a=2$, and its extension to general $a$ is our goal. The travel coordinate forces points with sufficiently small gradient to lie at distance of order $\Delta/\eps$ from initialization, whereas the activation function keeps the noisy chain inactive near initialization and allows it to become active only where the distance-dependent variance budget is sufficiently large. Note that, this is a geometric requirement, as an unrestricted algorithm may jump to a distant point in one query. The established oracle complexity comes from learning the chain coordinates, rather than from the number of steps needed to travel that distance. Note that, the proof strategies in \cite{fazla2026lower} are mostly built on top of \cite{arjevani2023lower}.

In our extension to BG-$a$, we replace PASTA's quadratic activation variance budget by $B^2(D/2)^a+G^2$ and recalibrate the Bernoulli reveal probability $p$. The chain length remains of order $L\Delta/\eps^2$, while each coordinate revelation requires order $1/p$ calls, with
$$
\frac1p=1+\Theta_a\!\left(\frac{G^2+B^2(\Delta/\eps)^a}{\eps^2}\right).
$$
Multiplying these two scales gives the claimed rate. We first establish this mechanism for zero-respecting algorithms, then give the compressed random-rotation construction and the random-stopping argument needed for unrestricted randomized algorithms.

\textbf{Stage 1: Construction and lower bound for zero-respecting algorithms.}
Let $D=\Delta/(4\eps)$ and define $f_0(0)=0$ with
\begin{equation}\label{eq:pasta-travel}
f_0'(u)=\begin{cases}
-2\eps,&u\le D,\\
-2\eps+\frac L2(u-D),&D<u<D+4\eps/L,\\
0,&u\ge D+4\eps/L.
\end{cases}
\end{equation}
Writing $f_0^{\inf}:=\inf_u f_0(u)$, integration gives $f_0^{\inf}=-\Delta/2-4\eps^2/L$. Next, we define the twice continuously differentiable activation
\begin{equation}\label{eq:pasta-activation}
\phi(u)=\begin{cases}
0,&u\le D/2,\\
10t^3-15t^4+6t^5,&D/2<u<D,\quad t=2u/D-1,\\
1,&u\ge D,
\end{cases}
\end{equation}
which satisfies $0\le\phi\le1$, $\phi'\ge0$, and
$$
\norm{\phi'}_\infty=\frac{15\eps}{\Delta},\qquad
\norm{\phi''}_\infty=\frac{640\eps^2}{\sqrt3\Delta^2}.
$$

Use the chain $\bar f_T$ and oracle $\bar g_T$ from Definitions~\ref{def:lower_chain} and~\ref{def:lower_masked_oracle}, with $M_T=\sup_w\bar f_T(w)$. Lemmas~\ref{lem:lower_chain_properties}--\ref{lem:lower_masked_properties} provide the smoothness, range, gradient, and variance bounds used below.

Using the same scaling as in \cite{fazla2026lower}, we set
\begin{equation}\label{eq:pasta-scaled}
f_{\mathrm{scaled}}(y)=\frac{L\lambda^2}{2\ell_1}
[\bar f_T(y/\lambda)-M_T],\qquad
\lambda=\frac{4\ell_1\eps}{L},\qquad
T=\left\lfloor\frac{L\Delta}{768\ell_1\eps^2}\right\rfloor.
\end{equation}
Then $-\Delta/4\le f_{\mathrm{scaled}}\le0$,
$\operatorname{Lip}(\nabla f_{\mathrm{scaled}})\le L/2$, and
$\nabla f_{\mathrm{scaled}}(y)=2\eps\nabla\bar f_T(y/\lambda)$.
Hence, subtracting $M_T$ makes the scaled chain nonpositive. Consequently, since $\phi'\ge0,$ the activation contribution $\phi'(u)f_{\mathrm{scaled}}(y)$ cannot cancel the negative travel-coordinate gradient.
Defining
\begin{equation}\label{eq:pasta-objective}
f(u,y)=f_0(u)+\phi(u)f_{\mathrm{scaled}}(y),\qquad x_0=(0,0),
\end{equation}
and writing $x=(u,y)$ and $f^{\inf}:=\inf_x f(x)$, we get $f(x_0)=0$ and
$$
f(x_0)-f^{\inf}\le\frac{3\Delta}{4}+\frac{4\eps^2}{L}\le\Delta.
$$

The block-diagonal part of $\nabla^2f$ has operator norm at most
$L/2+\norm{\phi''}_\infty\norm{f_{\mathrm{scaled}}}_\infty$.
The off-diagonal part has operator norm $\norm{\phi'\nabla f_{\mathrm{scaled}}}$, and
$\norm{\nabla f_{\mathrm{scaled}}}\le46\eps\sqrt T$.
Consequently,
\begin{equation}\label{eq:pasta-smoothness}
\norm{\nabla^2f}_{\mathrm{op}}
\le\frac L2+\frac{160\eps^2}{\sqrt3\Delta}
+\frac{690\eps\sqrt L}{\sqrt{768\ell_1\Delta}}
\le L.
\end{equation}
Indeed, $\eps^2\le L\Delta/(1536\ell_1)$, which follows from \eqref{eq:small}, bounds each of the last two terms by $L/4$. The gradient is continuous at the junctions of $f_0'$, so integration along line segments extends the bound to global $L$-smoothness.

Next, observe that the composite gradient is
$$
\nabla f(u,y)=\begin{pmatrix}
f_0'(u)+\phi'(u)f_{\mathrm{scaled}}(y)\\
\phi(u)\nabla f_{\mathrm{scaled}}(y)
\end{pmatrix}.
$$
We consider two cases to identify necessary conditions for $\norm{\nabla f(u,y)}<2\eps$.

First, if $u\le D$, then $f_0'(u)=-2\eps$. Since $\phi'(u)\ge0$ and $f_{\mathrm{scaled}}(y)\le0$, we have
$$
\partial_u f(u,y)
=-2\eps+\phi'(u)f_{\mathrm{scaled}}(y)
\le-2\eps.
$$
Therefore, $\norm{\nabla f(u,y)}\ge|\partial_u f(u,y)|\ge2\eps$. Thus, a gradient norm strictly below $2\eps$ requires $u>D$.

Second, suppose $u>D$ but $\operatorname{prog}_1(y/\lambda)<T$. Then $\phi(u)=1$, and the scaling in \eqref{eq:pasta-scaled} together with the gradient lower bound in Lemma~\ref{lem:lower_chain_properties} gives
$$
\begin{aligned}
\norm{\nabla f(u,y)}
&\ge\norm{\nabla_y f(u,y)}\\
&=\norm{\nabla f_{\mathrm{scaled}}(y)}\\
&=2\eps\norm{\nabla\bar f_T(y/\lambda)}\\
&\ge2\eps\norm{\nabla\bar f_T(y/\lambda)}_\infty
>2\eps.
\end{aligned}
$$
This also rules out a gradient norm strictly below $2\eps$. Since $\operatorname{prog}_1(y/\lambda)\in\{0,\ldots,T\}$, the two cases imply
\begin{equation}\label{eq:pasta-stationarity}
\norm{\nabla f(u,y)}<2\eps
\quad\Longrightarrow\quad u>D\ \text{and}\ \operatorname{prog}_1(y/\lambda)=T.
\end{equation}

We now choose a constant variance budget that is admissible wherever the chain is active. Recall that $x=(u,y)$, $x_0=(0,0)$, and $\phi(u)=0$ for $u\le D/2$. Thus, whenever $\phi(u)>0$,
$$
u>D/2>0,\qquad
\norm{x-x_0}=\sqrt{u^2+\norm{y}^2}\ge u>D/2.
$$
Since $r\mapsto r^a$ is nondecreasing for $a\ge0$, the BG-$a$ variance bound throughout this active region satisfies
$$
B^2\norm{x-x_0}^a+G^2\ge B^2(D/2)^a+G^2.
$$
We therefore define the constant budget
\begin{equation}\label{eq:pasta-variance-budget}
\widetilde\sigma_a^2 :=B^2(D/2)^a+G^2
=B^2\left(\frac{\Delta}{8\eps}\right)^a+G^2,
\end{equation}
where we use $D=\Delta/(4\eps)$ in the last equality. Note that, this budget is a permissible upper bound for the noise we introduce on the active region, it is not a lower bound on the oracle's actual variance. Thus, we construct the oracle below so that its variance is at most $\phi(u)^2\widetilde\sigma_a^2$. Since $0\le\phi\le1$, the factor $\phi(u)^2$ makes the noise vanish on the inactive region and keeps it within the chosen budget on the active region.

As the next step, we keep the travel-coordinate component of the gradient exact and introduce randomness only through the chain estimator $\bar g_T$. Its Bernoulli seed $\xi\sim\mathrm{Bernoulli}(p)$ permits the next unrevealed chain coordinate to be revealed only when $\xi=1$. The factor $\xi/p$ in the masked estimator preserves its expectation, while a smaller $p$ increases its variance through the factor $(1-p)/p$ in \eqref{eq:chainvariance}. We define the composite oracle and choose $p$ to match this variance bound to the budget above:
\begin{equation}\label{eq:pasta-oracle}
\nabla f(u,y;\xi)=\begin{pmatrix}
f_0'(u)+\phi'(u)f_{\mathrm{scaled}}(y)\\
2\eps\phi(u)\bar g_T(y/\lambda;\xi)
\end{pmatrix},\qquad
p=\frac{4\varsigma^2\eps^2}
{4\varsigma^2\eps^2+\widetilde\sigma_a^2}.
\end{equation}
To verify conditional unbiasedness, let $x_t=(u_t,y_t)$ be a query measurable with respect to $\cF_t$, and draw $\xi_t\sim\mathrm{Bernoulli}(p)$ independently of $\cF_t$. Conditioned on this history, the query is fixed, so \eqref{eq:chainvariance} gives
$$
\E_t[\bar g_T(y_t/\lambda;\xi_t)]
=\nabla\bar f_T(y_t/\lambda).
$$
Consequently, using $\nabla f_{\mathrm{scaled}}(y_t)=2\eps\nabla\bar f_T(y_t/\lambda)$,
$$
\begin{aligned}
\E_t[\nabla f(x_t;\xi_t)]
&=\begin{pmatrix}
f_0'(u_t)+\phi'(u_t)f_{\mathrm{scaled}}(y_t)\\
2\eps\phi(u_t)\nabla\bar f_T(y_t/\lambda)
\end{pmatrix}\\
&=\nabla f(x_t).
\end{aligned}
$$
The oracle error has a zero travel-coordinate component and can be written explicitly as
$$
e_t:=\nabla f(x_t;\xi_t)-\nabla f(x_t)
=\begin{pmatrix}
0\\
2\eps\phi(u_t)\bigl[\bar g_T(y_t/\lambda;\xi_t)-\nabla\bar f_T(y_t/\lambda)\bigr]
\end{pmatrix}.
$$
Taking its conditional second moment and applying \eqref{eq:chainvariance} yields
$$
\begin{aligned}
\E_t\norm{e_t}^2
&=4\eps^2\phi(u_t)^2
\E_t\norm{\bar g_T(y_t/\lambda;\xi_t)-\nabla\bar f_T(y_t/\lambda)}^2\\
&\le4\eps^2\phi(u_t)^2\varsigma^2\frac{1-p}{p}.
\end{aligned}
$$
On the other hand, the choice of $p$ in \eqref{eq:pasta-oracle} lies in $(0,1]$ and satisfies
$$
\frac{1-p}{p}
=\frac{\widetilde\sigma_a^2}{4\varsigma^2\eps^2}.
$$
Thus, the conditional variance satisfies the desired bound
$$
\E_t\norm{\nabla f(x_t;\xi_t)-\nabla f(x_t)}^2
\le\phi(u_t)^2\widetilde\sigma_a^2.
$$
For $u_t\le D/2$, we have $\phi(u_t)=0$, so the conditional variance is zero. For $u_t>D/2$, the active-region bound above gives
$$
\phi(u_t)^2\widetilde\sigma_a^2
\le\widetilde\sigma_a^2
=B^2(D/2)^a+G^2
\le B^2\norm{x_t-x_0}^a+G^2.
$$
Hence, conditional unbiasedness and the BG-$a$ variance bound hold at every query and for every $a\ge0$, including those in the activation interval $D/2<u_t<D$. To recover the BG-$0$ case, setting $a=2$ in \eqref{eq:pasta-variance-budget} gives $\widetilde\sigma_2^2=B^2\Delta^2/(64\eps^2)+G^2$. Substituting this expression into $p$ and multiplying its numerator and denominator by $64\eps^2$ gives
$$
p=\frac{256\varsigma^2\eps^4}
{B^2\Delta^2+64G^2\eps^2+256\varsigma^2\eps^4},
$$
recovering the choice in \cite{fazla2026lower}.

We now translate the stationarity requirement into a lower bound on the number of oracle calls. Consider a zero-respecting algorithm as in Definition~\ref{def:lower_zero_respecting}, with output written as $\widehat x=(\widehat u,\widehat y)$, and define the event $A:=\{\norm{\nabla f(\widehat x)}<2\eps\}.$
By \eqref{eq:criterion} and Markov's inequality, we have
$$
\Pr(A^c)
=\Pr\bigl(\norm{\nabla f(\widehat x)}\ge2\eps\bigr)
\le\frac{\E\norm{\nabla f(\widehat x)}}{2\eps}
\le\frac12,
$$
so $\Pr(A)\ge1/2$. Within the event $A$, \eqref{eq:pasta-stationarity} requires $\operatorname{prog}_1(\widehat y/\lambda)=T$, and in particular the $T$-th chain coordinate of the output must be nonzero. The zero-respecting restriction requires this coordinate to have been revealed before the algorithm outputs it. Starting from $y_0=0$, Lemma~\ref{lem:lower_masked_properties} implies that the masked chain oracle can reveal at most one new chain coordinate per oracle call, and only when the Bernoulli seed equals one. Thus, reaching the $T$-th coordinate requires at least $T$ Bernoulli successes.

Next, we define $\tau$ as the number of oracle calls and denote the number of successes before termination by
$$
S_\tau:=\sum_{t=0}^{\tau-1}\xi_t.
$$
The preceding argument gives $S_\tau\ge T\mathbf1_A$, hence
$$
\E[S_\tau]\ge T\Pr(A)\ge T/2.
$$
Observing that the event $\{\tau>t\}$ means that call $t$ is made and is $\cF_t$-measurable and using independence, we obtain
$$
\begin{aligned}
\E[\mathbf1_{\{\tau>t\}}\xi_t]
&=\E\bigl[\mathbf1_{\{\tau>t\}}\E[\xi_t\mid\cF_t]\bigr]\\
&=p\Pr(\tau>t).
\end{aligned}
$$
Next, using  Tonelli's theorem and the tail-sum identity for $\tau$ yields
$$
\begin{aligned}
\E[S_\tau]
&=\E\bigl[\sum_{t\ge0}\mathbf1_{\{\tau>t\}}\xi_t\bigr]\\
&=\sum_{t\ge0}\E[\mathbf1_{\{\tau>t\}}\xi_t]\\
&=p\sum_{t\ge0}\Pr(\tau>t)
=p\E[\tau].
\end{aligned}
$$
These identities also hold for infinite expectations, in which case the claimed lower bound is immediate. Otherwise, combining them with $\E[S_\tau]\ge T/2$ and dividing by $p>0$ gives $\E[\tau]\ge T/(2p)$.

For convenience, we weaken this to $\E[\tau]\ge T/(4p)$. The accuracy restriction \eqref{eq:small} ensures $L\Delta/(768\ell_1\eps^2)\ge2$, so the definition of $T$ and $\lfloor r\rfloor\ge r/2$ for $r\ge2$ give
$$
T\ge\frac{L\Delta}{1536\ell_1\eps^2}.
$$
Furthermore, \eqref{eq:pasta-oracle} and \eqref{eq:pasta-variance-budget} imply
$$
\frac1p
=1+\frac{\widetilde\sigma_a^2}{4\varsigma^2\eps^2}
=1+\frac{G^2+B^2(\Delta/(8\eps))^a}{4\varsigma^2\eps^2}.
$$
Substituting both bounds yields
\begin{equation}\label{eq:pasta-cost}
\E[\tau]\ge\frac{T}{4p}
\ge\frac{L\Delta}{6144\ell_1\eps^2}
\left[1+\frac{G^2+B^2(\Delta/(8\eps))^a}{4\varsigma^2\eps^2}\right].
\end{equation}

Hence, the expected number of oracle calls is lower bounded by the $\eps^{-6}$ growth term at $a=2,$ which becomes $\eps^{-(4+a)}$ for general $a$.

\textbf{Stage 2: Applying the reduction to unrestricted randomized algorithms.} Next, we extend the analysis to unrestricted randomized algorithms. 

Lemma~\ref{lem:bg_a_unrestricted_reduction} is built on an admissible activated-chain instance for unrestricted randomized algorithms, including unbounded queries, arbitrary output points, and random stopping. It combines the travel-coordinate and activation mechanism of ~\cite{fazla2026lower} with the compressed random-rotation construction of ~\cite{arjevani2023lower}. For its construction, we rely on a new chain length $T_\star$ and reveal probability $p_\star$, which replace the parameters in the first stage and are given as
$$
T_\star=\left\lfloor\frac{L\Delta}{12288\ell\eps^2}\right\rfloor,
\qquad
p_\star=\frac{64\varsigma^2\eps^2}
{64\varsigma^2\eps^2+G^2+B^2(\Delta/(16\eps))^a}.
$$

Equation \ref{eq:small} ensures that the expression inside the floor defining $T_\star$ is at least eight, which gives
$$
T_\star\ge\frac{L\Delta}{24576\ell\eps^2},\qquad
\frac1{p_\star}=1+\frac{G^2+B^2(\Delta/(16\eps))^a}{64\varsigma^2\eps^2}.
$$
Substituting these expressions into the inequality $\sup_{(f,g)}\E[\tau]\ge T_\star/(32p_\star),$ obtained in Lemma \ref{lem:bg_a_unrestricted_reduction}, yields
$$
\sup_{(f,g)}\E[\tau]
\ge\frac{L\Delta}{786432\ell\eps^2}
\left[1+\frac{G^2+B^2(\Delta/(16\eps))^a}{64\varsigma^2\eps^2}\right],
$$
which proves \eqref{eq:explicit} and \eqref{eq:rate}. Finally, Cauchy--Schwarz gives
$\E[\norm{\nabla f(\widehat x)}]\le(\E[\norm{\nabla f(\widehat x)}^2])^{1/2}$,
so the same bound applies to the squared gradient guarantee, which we utilize in our upper bound proofs.
\end{proof}

\subsection{Reduction to unrestricted randomized algorithms}\label{app:bg_a_unrestricted_reduction}

We first state the random rotation result used in the reduction. This is the single-query specialization of Lemma 6 of \cite{arjevani2023lower}, whose oracle returns the exact objective value together with the stochastic gradient. We use this version because the activated construction below requires the value of the rotated chain in order to compute the first, travel-coordinate component of the stochastic gradient oracle.

\begin{lemma}[Lemma 6 of \cite{arjevani2023lower} restated]\label{lem:lower_rotated_chain}
Let $T\ge1$, $p\in(0,1]$, and $\delta\in(0,1)$. Suppose
$$
d\ge\left\lceil18\cdot230^2\frac{T^2}{p}
\log\!\left(\frac{2T^2}{p\delta}\right)\right\rceil,
$$
and let $U\in\R^{d\times T}$ be uniformly distributed over matrices with orthonormal columns, independently of the algorithm's internal randomness and oracle seeds. We define
\begin{align*}
\rho(z)&=z/\sqrt{1+\norm z^2/R^2},\qquad R=230\sqrt T,\qquad J(z)=D\rho(z),\\
h_U(z)&=\bar f_T(U^\top\rho(z)),\qquad q(z)=\norm z^2/10,\qquad H_U=h_U+q,\\
v_U(z;\xi)&=J(z)^\top U\bar g_T(U^\top\rho(z);\xi)+z/5,
\end{align*}
where $J(z)=D\rho(z)$ denotes the Jacobian of $\rho$ and $\xi\sim\mathrm{Bernoulli}(p)$. The algorithm accesses the unknown instance through the responses $(H_U(z),v_U(z;\xi))$. For any adaptive randomized algorithm making one query per call, with no bound on the query norms, and with probability at least $1-\delta$,
$$
\norm{\nabla H_U(z_j)}\ge\frac12
\quad\text{for every integer }1\le j\le\frac{T-\log(2/\delta)}{2p}.
$$
\end{lemma}

\begin{lemma}[Unrestricted algorithm reduction of the activated BG-$a$ chain]\label{lem:bg_a_unrestricted_reduction}
Fix $L,\Delta>0$, $B,G\ge0$, and a finite $a\ge0$. Let $\ell=155$, $\varsigma=23$, and $0<\eps^2\le L\Delta/(98304\ell)$. Define
$$
T=\left\lfloor\frac{L\Delta}{12288\ell\eps^2}\right\rfloor,
\qquad
p=\frac{64\varsigma^2\eps^2}
{64\varsigma^2\eps^2+G^2+B^2(\Delta/(16\eps))^a}.
$$
For any integer $d$ satisfying
\begin{equation}\label{eq:dimension}
d\ge\left\lceil18\cdot230^2\frac{T^2}{p}
\log\!\left(\frac{8T^2}{p}\right)\right\rceil,
\end{equation}
there is a family of lower bounded, globally $L$-smooth objectives $f_U:\R^{d+1}\to\R$, indexed by matrices $U\in\R^{d\times T}$ with orthonormal columns, and unbiased vector oracles $g_U(x;\xi)=\nabla f_U(x;\xi)$ such that, for $x_0=0$ and $f_U^{\inf}:=\inf_x f_U(x)$,
$$
f_U(x_0)-f_U^{\inf}\le\Delta,\qquad
\E_\xi\norm{g_U(x;\xi)-\nabla f_U(x)}^2
\le B^2\norm{x-x_0}^a+G^2
\quad\text{for every }x.
$$
Every adaptive randomized algorithm that returns $\widehat x$ at a stopping time $\tau$ and satisfies $\E\norm{\nabla f_U(\widehat x)}\le \eps$ for every function in this family must obey
$$
\sup_U \E[\tau \mid U] \ge \frac{T}{32p}.
$$
\end{lemma}

\begin{proof}
The random rotation makes the chain directions inaccessible to the algorithm. Compression keeps the input to the chain bounded, even for unbounded queries. Finally, the quadratic penalty ensures that large-norm queries cannot exploit the compression to produce small gradients. 

Now, we use $U$, $\rho$, $J$, $h_U$, $q$, $H_U$, and $v_U$ from Lemma~\ref{lem:lower_rotated_chain} with the parameters $T,p$ specified above.
Here $\norm J_{\mathrm{op}}\le1$ and $\operatorname{Lip}(J)\le3/R$, so we have
$\operatorname{Lip}(\nabla h_U)\le152+69/230<\ell$,
$\norm{\nabla h_U}\le23\sqrt T$, and $v_U$ is unbiased for $\nabla H_U$ with variance at most $\varsigma^2(1-p)/p$.
Taking $\delta=1/4$ in Lemma~\ref{lem:lower_rotated_chain}, we obtain the dimension requirement \eqref{eq:dimension}. Hence, with probability at least $3/4$, any adaptive randomized query sequence satisfies
\begin{equation}\label{eq:primitive}
\norm{\nabla H_U(z_j)}\ge\frac12
\quad\text{for every integer }1\le j\le\frac{T-\log8}{2p}.
\end{equation}

For the activated construction, we use the following scales
\begin{align}
D&=\frac{\Delta}{8\eps},\qquad
s=\frac{16\ell\eps}{L},\qquad \kappa=\frac{Ls^2}{2\ell},\notag\\
T&=\left\lfloor\frac{L\Delta}{12288\ell\eps^2}\right\rfloor,
\qquad V_a=B^2(D/2)^a+G^2,\qquad
p=\frac{64\varsigma^2\eps^2}{64\varsigma^2\eps^2+V_a}.
\label{eq:Tp}
\end{align}

Now, in a similar way to \eqref{eq:pasta-travel}--\eqref{eq:pasta-activation}, with $\eps$ replaced by $2\eps$, we set
\begin{equation}\label{eq:objective}
A_U(y)=\kappa[h_U(y/s)-M_T],\qquad
f_U(u,y)=f_0(u)+\phi(u)A_U(y)+\kappa q(y/s).
\end{equation}
The same shifted chain bounds give $-\Delta/4\le A_U\le0$,
$\norm{\nabla A_U}\le184\eps\sqrt T$, and
$\operatorname{Lip}(\nabla A_U)\le L/2$. The gap is at most
$3\Delta/4+16\eps^2/L\le\Delta$, and the Hessian bound of the block becomes
$$
\norm{\nabla^2f_U}_{\mathrm{op}}
\le\frac L2+\frac{L}{10\ell}
+\frac{640\eps^2}{\sqrt3\Delta}
+\frac{5520\eps^2\sqrt T}{\Delta}
\le\frac{7L}{8}.
$$
The last three terms are each at most $L/8$ by \eqref{eq:small}, and continuity at the travel-function junctions again gives global smoothness. For $u\le D$, we have $\partial_uf_U\le-4\eps$; and for $u>D$, we get $\nabla_yf_U=8\eps\nabla H_U(y/s)$.
Thus \eqref{eq:primitive} forces $\norm{\nabla f_U}\ge4\eps$ regardless of $u$. Denoting $z=y/s$, we get
\begin{equation}\label{eq:oracleconstruction}
\nabla f_U(u,y;\xi)=\begin{pmatrix}
f_0'(u)+\phi'(u)A_U(y)\\
8\eps\{\phi(u)[v_U(z;\xi)-z/5]+z/5\}
\end{pmatrix},
\end{equation}
which is conditionally unbiased with conditional variance at most
$64\eps^2\phi(u)^2\varsigma^2(1-p)/p=\phi(u)^2V_a$. Note that both the inactive and active cases used above satisfy \eqref{eq:model} at every point. As $A_U(y)=\kappa[H_U(z)-q(z)-M_T]$, each composite call can be simulated with one call to $(H_U,v_U)$ at $z=y/s$. Hence the first coordinate reveals no additional information about $U$.

It remains to turn the query bound in \eqref{eq:primitive} into a lower bound on the expected stopping time. Setting $m=\lfloor T/(8p)\rfloor$, the accuracy restriction gives $T\ge8$, and $p\le1$, so $T/(8p)\ge1$. Therefore,
$$
m\ge\frac{T}{16p},\qquad
m+1\le\frac{T}{8p}+1\le\frac{T}{4p}
\le\frac{T-\log8}{2p},
$$
where we use $T\ge8>2\log8$ in the last inequality. Thus the bound in \eqref{eq:primitive} covers $m$ oracle calls with an additional query at the output point.

First, lets consider any algorithm that makes at most $m$ calls and returns $\widehat x=(\widehat u,\widehat y)$, and simulate these calls using the oracle $(H_U,v_U)$ as described above. If the algorithm stops before $m$ calls, we keep its output fixed and make the remaining queries at the origin, ignoring their responses. We then query $\widehat y/s$ as query $m+1$. This final query is used only to apply \eqref{eq:primitive} and its response does not change the output. With probability at least $3/4$, we get
$$
\norm{\nabla H_U(\widehat y/s)}\ge\frac12.
$$
If $\widehat u\le D$, the travel-coordinate derivative already has magnitude at least $4\eps$. If $\widehat u>D$, then
$$
\norm{\nabla f_U(\widehat x)}
\ge\norm{\nabla_y f_U(\widehat x)}
=8\eps\norm{\nabla H_U(\widehat y/s)}
\ge4\eps.
$$
Hence every algorithm with a budget of $m$ calls satisfies
$$
\Pr\bigl(\norm{\nabla f_U(\widehat x)}\ge4\eps\bigr)\ge\frac34.
$$

Now let $\mathcal A$ be the algorithm in Lemma \ref{lem:bg_a_unrestricted_reduction}, with stopping time $\tau$ and output $\widehat x$. Its guarantee holds for every fixed $U$, so averaging over $U$ gives $\E\norm{\nabla f_U(\widehat x)}\le\eps$. Markov's inequality then yields
$$
\Pr\bigl(\norm{\nabla f_U(\widehat x)}\ge4\eps\bigr)
\le\frac{\E\norm{\nabla f_U(\widehat x)}}{4\eps}
\le\frac14.
$$
Next, we define a truncated algorithm $\mathcal A_m$ that runs $\mathcal A$ for at most $m$ calls. If $\mathcal A$ stops within this budget, $\mathcal A_m$ returns the same output. Otherwise, it returns $x_0$. Denoting $\widehat x_m$ as this truncated algorithm's output, we run the two algorithms with the same $U$, oracle seeds, and internal randomness. Their outputs agree whenever $\tau\le m$. Consequently,
$$
\begin{aligned}
\Pr\bigl(\norm{\nabla f_U(\widehat x_m)}\ge4\eps\bigr)
&\le\Pr\bigl(\norm{\nabla f_U(\widehat x)}\ge4\eps\bigr)
+\Pr(\tau>m)\\
&\le\frac14+\Pr(\tau>m).
\end{aligned}
$$
Since $\mathcal A_m$ uses at most $m$ calls, its probability on the left is at least $3/4$. Thus
$$
\Pr(\tau>m)\ge\frac34-\frac14=\frac12.
$$
Finally, $\tau\ge m\mathbf1_{\{\tau>m\}}$, and the supremum over fixed rotations is at least the average over $U$. Therefore,
\begin{equation}\label{eq:expected}
\sup_U\E[\tau\mid U]
\ge\E[\tau]
\ge m\Pr(\tau>m)
\ge\frac m2
\ge\frac{T}{32p},
\end{equation}
which proves the expected SFO lower bound and completes the lemma.
\end{proof}

\subsection{Interpretation and relation to the upper bound}
Based on our lower bound proofs from the previous part, the chain length is $T=\Theta(L\Delta/\eps^2)$, while the inverse reveal probability is
\begin{equation}
\frac1p=1+\Theta_a\!\left(\frac{G^2+B^2(\Delta/\eps)^a}{\eps^2}\right).
\end{equation}
Therefore, the expected number of oracle calls is at least of order $T/p$. For varying values of $a$, this gives:

\begin{center}
\begin{tabular}{lll}
\toprule
Distance Exponent & Variance Bound & Lower Bound\\
\midrule
$a=0$ & $B^2+G^2$ & $\Omega(\eps^{-4})$\\
$0<a<2$ & $B^2\norm{x-x_0}^a+G^2$ & $\Omega_a(\eps^{-(4+a)})$\\
$a=2$ & $B^2\norm{x-x_0}^2+G^2$ & $\Omega(\eps^{-6})$\\
\bottomrule
\end{tabular}
\end{center}
Note that, if $B=0$ and $G>0$, the BG-$a$ noise model reduces to the more restrictive bounded variance case, and the lower bound becomes $\Omega(\eps^{-4})$, recovering \cite{arjevani2023lower}. Similarly, if $B=G=0$, only the deterministic $\epsilon^{-2}$ term remains, and the lower bound reduces to the deterministic case. The additional factor $\eps^{-a}$ to the lower bounds obtained in \cite{fazla2026lower} is the distance-dependent term and comes from the allowable variance at distance $\Theta(\Delta/\eps)$ from initialization. For $0\le a\le2$ and $B>0$, the lower bound matches the $\eps$-dependence of the expected SFO upper bounds for dynamically batched SGD and SGDM, obtained in Theorems~\ref{thm:main_sgd} and~\ref{thm:main_sgdm}, for fixed problem parameters, target variance $\sigma^2$, and momentum coefficient $\beta$. 

\section{Convergence of SGD}\label{sec:convergence_of_sgd}

\begin{theorem}[Convergence of SGD with Dynamic Batching]\label{thm:sgd_conv} Suppose Assumptions \ref{as:l_smooth} and \ref{as:bg_a_oracle} hold with $0\le a\le2$. Under a fixed target variance $\sigma^2>0$ and a target accuracy $\eps > 0$ satisfying $\eps^2\le\min\{2\sigma^2,L\Delta\}$, denote the batch size $b_t$, step size $\eta$, and total iterations $T$ as
\begin{equation}\label{eq:sgd_convergence_parameters}
b_t=\max\!\left\{1,\left\lceil\frac{B^2\norm{x_t-x_0}^a+G^2}{\sigma^2}\right\rceil\right\},\qquad 
\eta=\frac{\eps^2}{2L\sigma^2},\qquad
T=\left\lceil\frac{4\Delta}{\eta\eps^2}\right\rceil.
\end{equation}
For $t=0,\ldots,T-1$, form the minibatch gradient
$g_t=\frac1{b_t}\sum_{i=1}^{b_t}\nabla f(x_t;\xi_{t,i})$
from fresh conditionally independent samples and update $x_{t+1}=x_t-\eta g_t.$ Then,
\begin{equation}
\frac1T\sum_{t=0}^{T-1}\E\norm{\nabla f(x_t)}^2\le\eps^2,
\end{equation}
and the expected total stochastic first-order (SFO) oracle complexity satisfies
\begin{equation}
\E[\text{SFO}] = \E\left[\sum_{t=0}^{T-1}b_t\right]
\le 10L\Delta\left[(\sigma^2+G^2)\eps^{-4}
+50^{a/2}B^2\Delta^a\eps^{-(4+a)}\right].
\end{equation}
In particular, for a fixed $\sigma^2$ independent of $\epsilon$, the expected SFO complexity is $\Ocal(\eps^{-(4+a)})$.
\end{theorem}

\begin{proof} Let $e_t:=g_t-\nabla f(x_t)$, and $z_{t,i}:=\nabla f(x_t;\xi_{t,i})-\nabla f(x_t)$. Note that the batch size $b_t$ is chosen \emph{before} sampling that batch. Hence, conditioned on $\cF_t$, $b_t$ is a fixed positive integer and $z_{t,i}$ are independent with conditional mean zero. Therefore, we have
\begin{align}\label{eq:cap}
\E_t[e_t] = 0, \qquad
\E_t\norm{e_t}^2 = \frac1{b_t^2}\sum_{i=1}^{b_t}\E_t\norm{z_{t,i}}^2
\le\frac{B^2\norm{x_t-x_0}^a+G^2}{b_t}
\le\sigma^2,
\end{align}
where the cross terms vanish by conditional independence and zero means.


\textbf{Step 1: Descent lemma.}
Using $L$-smoothness and with the SGD update iterates $x_t$, we get
\begin{equation}
f(x_{t+1})\le f(x_t)-\eta\langle\nabla f(x_t),g_t\rangle
+\frac{L\eta^2}{2}\norm{g_t}^2.
\end{equation}
Taking conditional expectation $\E_t[\cdot]$ and using \eqref{eq:cap}
\begin{equation}
\E_t\norm{g_t}^2
=\norm{\nabla f(x_t)}^2+\E_t\norm{e_t}^2
\le\norm{\nabla f(x_t)}^2+\sigma^2.
\end{equation}
Consequently, since $\eta\le1/L$ (as $\eps^2 \leq 2\sigma^2$),
\begin{align}
\E_t f(x_{t+1})
&\le f(x_t)-\eta\left(1-\frac{L\eta}{2}\right)\norm{\nabla f(x_t)}^2
+\frac{L\eta^2\sigma^2}{2}\notag\\
&\le f(x_t)-\frac\eta2\norm{\nabla f(x_t)}^2
+\frac{L\eta^2\sigma^2}{2}.\label{eq:descent}
\end{align}

\textbf{Step 2: Telescoping to obtain the stationarity criterion.} Taking total expectation of each side of \eqref{eq:descent}, summing over $t=0,\ldots,T-1$, and since $\E[f(x_T)]\ge f^{\inf},$ we get 
\begin{align}
\frac\eta2\sum_{t=0}^{T-1}\E\norm{\nabla f(x_t)}^2
&\le f(x_0)-\E[f(x_T)]+\frac{TL\eta^2\sigma^2}{2}\notag\\
&\le\Delta+\frac{TL\eta^2\sigma^2}{2}.\label{eq:telescope}
\end{align}
Dividing both sides by $T\eta/2$ and substituting the parameters from \eqref{eq:sgd_convergence_parameters}, we end up with
\begin{equation}\label{eq:accuracy}
\frac1T\sum_{t=0}^{T-1}\E\norm{\nabla f(x_t)}^2
\le\frac{2\Delta}{T\eta}+L\eta\sigma^2
\le\frac{\eps^2}{2}+\frac{\eps^2}{2}=\eps^2.
\end{equation}

\textbf{Step 3: SFO computation.} Telescoping the iterates $x_t$ gives:
\begin{equation}
x_t-x_0=-\eta\sum_{j=0}^{t-1}\nabla f(x_j)-\eta\sum_{j=0}^{t-1}e_j.
\end{equation}
Taking the norm squared into total expectation of both sides gives
\begin{align}
\E\norm{x_t-x_0}^2 = \E\norm{\eta\sum_{j=0}^{t-1}\nabla f(x_j)+\eta\sum_{j=0}^{t-1}e_j}^2
\end{align}
Applying the inequality $\norm{u+v}^2\le2\norm u^2+2\norm v^2$ and Cauchy--Schwarz on the gradient sum gives
\begin{equation}\label{eq:split}
\E\norm{x_t-x_0}^2
\le2\eta^2t\sum_{j=0}^{t-1}\E\norm{\nabla f(x_j)}^2
+2\eta^2\E\left\|\sum_{j=0}^{t-1}e_j\right\|^2.
\end{equation}
Now, notice that for $i<j$, $e_i$ is $\cF_j$-measurable, hence
\begin{equation}
\E[\langle e_i, e_j\rangle]=\E[\E[ \langle e_i, e_j \rangle\mid \cF_j]] = \E[\langle e_i, \E[e_j \mid \cF_j]\rangle]=0.
\end{equation}
Thus, martingale orthogonality and \eqref{eq:cap} imply
\begin{equation}
\E\left\|\sum_{j=0}^{t-1}e_j\right\|^2
=\sum_{j=0}^{t-1}\E\norm{e_j}^2\le t\sigma^2.
\end{equation}
Substituting this and \eqref{eq:telescope} into \eqref{eq:split}, and using $t\le T$, we obtain
\begin{align}
\E\norm{x_t-x_0}^2
&\le2\eta^2T\left(\frac{2\Delta}{\eta}+TL\eta\sigma^2\right)
+2\eta^2T\sigma^2\notag\\
&=4\eta T\Delta+2L\eta^3T^2\sigma^2+2\eta^2T\sigma^2=:D_T.\label{eq:radius}
\end{align}

\textbf{Step 4: $a$ power growth.}
For $0\le a\le2$, the map $u\mapsto u^{a/2}$ is concave on $[0,\infty)$. Hence, using Jensen's inequality gives
\begin{equation}\label{eq:sgd_power_growth_jensen}
\E\norm{x_t-x_0}^a
=\E\left[(\norm{x_t-x_0}^2)^{a/2}\right]
\le\left(\E\norm{x_t-x_0}^2\right)^{a/2}
\le (D_T)^{a/2}.
\end{equation}
Summing $b_t$, taking expectations and using $\max\{1,\lceil u\rceil\}\le1+u$ for $u\ge0$, we get
\begin{align}
\E \left[ \sum_{t=0}^{T-1} b_t\right]
&\le\sum_{t=0}^{T-1}\left[1+\frac{G^2}{\sigma^2}
+\frac{B^2}{\sigma^2}\E[\norm{x_t-x_0}^a]\right]\notag\\
&\le T\left(1+\frac{G^2}{\sigma^2}+\frac{B^2}{\sigma^2}(D_T)^{a/2}\right). \label{eq:exp_of_bt}
\end{align}

\textbf{Step 5: Substituting the explicit constants.}
Setting $\eps^2\le\min\{2\sigma^2,L\Delta\}$ and using the explicit constant values from Theorem \ref{thm:sgd_conv} gives us
\begin{equation}
\eta=\frac{\eps^2}{2L\sigma^2},\qquad
m:=\frac{4\Delta}{\eta\eps^2}=\frac{8L\sigma^2\Delta}{\eps^4}\ge4.
\end{equation}
Since $m\ge4$, the ceiling satisfies $T=\lceil m\rceil\le m+1\le(5/4)m$. Hence
\begin{equation}\label{eq:Kbound}
T\le\frac{10L\sigma^2\Delta}{\eps^4},\qquad
\eta T\le\frac{5\Delta}{\eps^2}.
\end{equation}
We now bound the three terms of $D_T$ (\eqref{eq:radius}) separately:
\begin{align}
4\eta T\Delta&\le\frac{20\Delta^2}{\eps^2},\\
2L\eta^3T^2\sigma^2
=2L\eta\sigma^2(\eta T)^2
&\le\eps^2\frac{25\Delta^2}{\eps^4}
=\frac{25\Delta^2}{\eps^2},\\
2\eta^2T\sigma^2
=2\eta\sigma^2(\eta T)
&\le\frac{\eps^2}{L}\frac{5\Delta}{\eps^2}
=\frac{5\Delta}{L}\le\frac{5\Delta^2}{\eps^2},
\end{align}
where the final inequality uses $\eps^2\le L\Delta$. Combining all three bounds, we get
\begin{equation}\label{eq:sharpD}
D_T\le\frac{50\Delta^2}{\eps^2}.
\end{equation}
Combining \eqref{eq:Kbound}, \eqref{eq:sharpD}, and \eqref{eq:exp_of_bt}, we arrive at
\begin{equation}\label{eq:sgd_explicitcost}
\E \left[ \text{SFO} \right] \le10L\Delta\left[(\sigma^2+G^2)\eps^{-4}
+50^{a/2}B^2\Delta^a\eps^{-(4+a)}\right],
\end{equation}
completing the proof.

\end{proof}

\textbf{Interpretation of \eqref{eq:sgd_explicitcost}.}
We demonstrate below the expected SFO based on the values of $a$, and discuss the relation to the convergence results in the literature.
\begin{center}
\begin{tabular}{@{}lll@{}}\toprule
Distance exponent & Variance upper bound & Expected SFO upper bound\\\midrule
$a=0$ & $B^2+G^2$ & $\Ocal(\eps^{-4})$\\
$0 < a < 2$ & $B^2\norm{x-x_0}^{a}+G^2$ & $\Ocal(\eps^{-(4+a)})$\\
$a=2$ & $B^2\norm{x-x_0}^{2}+G^2$ & $\Ocal(\eps^{-6})$\\\bottomrule
\end{tabular}
\end{center}

\textbf{Connecting bounded variance and BG-0 through $a$.}
The exponent $a$ connects the bounded variance setting of ordinary SGD with the BG-0 model of distance-dependent, potentially unbounded variance. At $a=0$, the convention $\norm{x-x_0}^0=1$ gives the uniform variance bound $B^2+G^2$. The batch size in \eqref{eq:sgd_convergence_parameters} is then constant throughout the run, and \eqref{eq:sgd_explicitcost} recovers the $\Ocal(\eps^{-4})$ SFO rate of SGD under bounded variance. Independently of $a$, setting $B=0$ also gives a uniform variance bound, now equal to $G^2$, and removes the distance-dependent term from the complexity bound.

At $a=2$, the oracle bound becomes $B^2\norm{x-x_0}^2+G^2$, which is the BG-0 model considered in \cite{fazla2026lower}. The expected SFO bound correspondingly becomes $\Ocal(\eps^{-6})$, recovering the smooth nonconvex BG-0 rate. For $0<a<2$, the rate $\Ocal(\eps^{-(4+a)})$ interpolates between these two endpoints: $a$ specifies the power of the allowed variance growth with distance from initialization. Note that, in our proof, the displacement scale $\Delta/\eps$ contributes a factor $(\Delta/\eps)^a$ to the expected batch cost, while the iteration budget remains $\Ocal(\eps^{-4})$ for fixed $\sigma^2$.


\subsection{Discussion on the Expected SFO bound when \texorpdfstring{$a>2$}{a>2}}
From our analysis in Theorem \ref{thm:sgd_conv}, we see that the descent and expected SFO analyses require different moments, given that the variance is of order $2$ while the distance norm is of order $a$. When $0\leq a\leq2$, meaning that the moment of the distance growth is of order smaller than or equal to the gradient noise variance, the mismatch in orders do not cause a problem, since we can utilize steps similar to \eqref{eq:sgd_power_growth_jensen}. However, when the order of the moment of the distance growth exceed that of the variance, we can no longer control it.

Formally, the variance bound in \eqref{eq:cap} controls the expected decrease in the objective, whereas the batch rule in \eqref{eq:sgd_convergence_parameters} gives
\begin{equation}\label{eq:batch_moment_requirement}
\frac{B^2\E\|x_t-x_0\|^a+G^2}{\sigma^2}
\le\E[b_t]
\le1+\frac{B^2\E\|x_t-x_0\|^a+G^2}{\sigma^2}.
\end{equation}
Thus, when $B>0$, finite expected batch cost requires a finite $a$-th moment of the displacement. In the proof of Theorem~\ref{thm:sgd_conv}, the second-moment bound \eqref{eq:radius} supplies this control through Jensen's inequality for $0<a\le2$. For $a>2$, a second-moment bound is no longer sufficient. To obtain the same $\Ocal(\eps^{-(4+a)})$ expected SFO bound with the prescribed $T=\Ocal(\eps^{-4})$ iterations, we need to establish
\begin{equation}\label{eq:required_displacement_moment}
\frac1T\sum_{t=0}^{T-1}\E\|x_t-x_0\|^a
=\Ocal\bigl((\Delta/\eps)^a\bigr).
\end{equation}
However, this construction may not be possible. As an example, the following construction shows why Assumptions \ref{as:l_smooth} and \ref{as:bg_a_oracle} alone cannot satisfy even the finiteness of this moment: the first SGD update can inherit an infinite $a$-th moment from a finite variance oracle.

\begin{proposition}[Expected SFO cost for $a>2$]\label{prop:obstruction}
For every fixed $a>2$ and target variance $\sigma^2>0$, there exist an objective and a stochastic gradient oracle satisfying Assumptions \ref{as:l_smooth} and \ref{as:bg_a_oracle} such that SGD with the batch rule in \eqref{eq:sgd_convergence_parameters} and any fixed step size $\eta>0$ satisfies $\E[b_1]=\infty$. Consequently, its expected total SFO cost is infinite for every $T\ge2$.
\end{proposition}

\begin{proof}
We demonstrate the proof by direct construction of such case. We set $d=1$, $x_0=0$, and $f(x)=(x-1)^2/2$, which gives $L=1$ and $\Delta=1/2$. Then, we choose a random variable $Z$ such that
\begin{equation}\label{eq:obstruction_noise_moments}
\E[Z]=0,\qquad \E[Z^2]=1,\qquad \E[|Z|^a]=\infty.
\end{equation}
For example, a symmetric Pareto variable with tail exponent $p\in(2,a)$ can be rescaled to have unit variance. We denote the stochastic signal $f(x;Z)=f(x)+Z\int_0^x\sqrt{\sigma^2+|s|^a}\,ds$, whose expectation gives the objective function $f(x)$ and its gradient can be written as
\begin{equation}
g(x;Z)=x-1+\sqrt{\sigma^2+|x|^a}\,Z,
\end{equation}
using an independent draw of $Z$ at each oracle call. Moreover, we have
\begin{equation}\label{eq:obstruction_variance}
\E_Z[g(x;Z)]=\nabla f(x),\qquad
\E_Z \norm{g(x;Z)-\nabla f(x)}^2=\sigma^2+|x|^a.
\end{equation}
Thus, the oracle satisfies Assumption \ref{as:bg_a_oracle} with $B=1$ and $G=\sigma$. Now, since the variance at $x_0=0$ equals the target variance, the initial batch size is $b_0=1$. The first update is therefore
\begin{equation}\label{eq:obstruction_first_update}
g_0=-1+\sigma Z_{0,1},\qquad x_1=\eta(1-\sigma Z_{0,1}),
\end{equation}
Where $Z_{0,1}$ is a single instantiation of $Z$ sampled at time $x_0$. From this update step, we can see that the second moment $\E[|x_1|^2]=\eta^2(1+\sigma^2)$ and is finite. However, its $a$-th moment is infinite. Indeed, $|1-\sigma Z_{0,1}|\ge\sigma|Z_{0,1}|/2$ whenever $|Z_{0,1}|\ge2/\sigma$, giving
\begin{equation}\label{eq:obstruction_displacement}
\E[|x_1|^a]
\ge\left(\frac{\eta\sigma}{2}\right)^a
\E\left[|Z_{0,1}|^a\mathbf{1}_{\{|Z_{0,1}|\ge2/\sigma\}}\right]
=\infty.
\end{equation}
The next batch size satisfies
\begin{equation}
b_1=\left\lceil1+\frac{|x_1|^a}{\sigma^2}\right\rceil
\ge1+\frac{|x_1|^a}{\sigma^2}.
\end{equation}
Hence \eqref{eq:obstruction_displacement} gives $\E[b_1]=\infty$, and
$\E[\sum_{t=0}^{T-1}b_t]\ge\E[b_1]=\infty$ for $T\ge2$.
\end{proof}

The obstruction is a mismatch between the moment controlled by the oracle assumption and the moment required by the batch rule. Variance control remains sufficient for the expected stationarity analysis, but it does not control the $a$-th displacement moment that determines expected SFO cost when $a>2$. Recovering the rate requires a quantitative bound such as \eqref{eq:required_displacement_moment} and finiteness alone would not establish its dependence on $\eps$.

Note that, such a quantitative bound can be obtained with new assumptions that bound the gradient noise moments of order $a$ when $a\geq2$, hence, matching the orders of noise growth and distance growth. This type of analysis, notation given below for convenience, requires further thought and new analysis. Interestingly, it resembles the heavy-tailed $p$-moment analysis common in the literature \citep{zhang2020adaptive}. We leave this interesting intersection of $p$-th moment growth and $a$-th moment distance growth for future work.

\begin{equation}\label{eq:pgrad_adist_growth}
\mathbb{E}_{\xi}
\left\|\nabla f(x;\xi)-\nabla f(x)\right\|^p
\le B^2\norm{x-x_0}^a+G^2.
\end{equation}

\section{Convergence of SGDM}\label{sec:convergence_of_sgdm}

\begin{theorem}[Convergence of SGDM with Dynamic Batching]\label{thm:sgdm_conv}\label{thm:main}
Suppose Assumptions \ref{as:l_smooth} and \ref{as:bg_a_oracle} hold with $0\le a\le2$. Fix the momentum coefficient $\beta\in[0,1)$ and a target variance $\sigma^2>0$, and define $\cB:=\frac{1-\beta}{1+\beta}.$ For a target accuracy $\eps > 0$ satisfying $\eps^2\le\min\{8\sigma^2(1-\beta^2),L\Delta\},$
denote the batch size $b_t$, step size $\eta$, and total iterations $T$ as
\begin{equation}\label{eq:sgdm_convergence_parameters}
\begin{aligned}
b_t&=\max\!\left\{1,\left\lceil
\cB\frac{B^2\norm{x_t-x_0}^a+G^2}{\sigma^2}
\right\rceil\right\},\\
\eta&=
\frac{\cB\eps^2}{16L\sigma^2},\qquad
T=\left\lceil\frac{16\Delta}{\eta\eps^2}\right\rceil.
\end{aligned}
\end{equation}
For $t=0,\ldots,T-1$, form the minibatch gradient
$g_t=\frac1{b_t}\sum_{i=1}^{b_t}\nabla f(x_t;\xi_{t,i})$
from fresh conditionally independent samples and update
\begin{equation}\label{eq:sgdm_update}
m_{-1}=0,\qquad m_t=\beta m_{t-1}+(1-\beta)g_t,\qquad
x_{t+1}=x_t-\eta m_t.
\end{equation}
Then
\begin{equation}\label{eq:sgdm_accuracy}
\frac1T\sum_{t=0}^{T-1}\E\norm{\nabla f(x_t)}^2\le\eps^2,
\end{equation}
and the expected total stochastic first-order (SFO) oracle complexity satisfies
\begin{equation}\label{eq:sgdm_explicitcost_}
\E[\text{SFO}] = \E\left[\sum_{t=0}^{T-1}b_t\right]
\le320L\Delta\left[
\left(\frac{\sigma^2}{\cB}+G^2\right)\eps^{-4}
+725^{a/2}B^2\Delta^a\eps^{-(4+a)}\right].
\end{equation}
In particular, for fixed $\beta$ and $\sigma^2$ independent of $\eps$, the expected SFO complexity is $\Ocal(\eps^{-(4+a)})$.
\end{theorem}

\begin{proof}
Let $e_t:=g_t-\nabla f(x_t)$ and
$z_{t,i}:=\nabla f(x_t;\xi_{t,i})-\nabla f(x_t)$.
The history $\cF_t$ contains all samples revealed before iteration $t$, therefore, $x_t$, $m_{t-1}$, and $b_t$ are $\cF_t$-measurable. We choose the batch size before sampling, and the errors $z_{t,i}$ are independent with conditional mean zero. Therefore,
\begin{align}\label{eq:sgdm_cap}
\E_t[e_t] = 0, \qquad
\E_t\norm{e_t}^2 =\frac1{b_t^2}\sum_{i=1}^{b_t}\E_t\norm{z_{t,i}}^2
\le\frac{B^2\norm{x_t-x_0}^a+G^2}{b_t}
\le\frac{\sigma^2}{\cB},
\end{align}
where the cross terms vanish due to conditional independence and zero means. The factor $\cB$ corresponds to the variance reduction of the momentum filter. Indeed, the filtered noise
$r_t:=(1-\beta)\sum_{j=0}^t\beta^{t-j}e_j$
satisfies, by martingale orthogonality
$$
\E\norm{r_t}^2
=(1-\beta)^2\sum_{j=0}^t\beta^{2(t-j)}\E\norm{e_j}^2
\le\frac{(1-\beta)^2}{1-\beta^2}\frac{\sigma^2}{\cB}
=\sigma^2.
$$
This controls the filtered noise component.

\textbf{Step 1: Descent lemma.}
To handle the components introduced by momentum, we define the auxiliary iterate
$$
\widetilde x_t:=x_t-\frac{\eta\beta}{1-\beta}m_{t-1},
$$
where $\widetilde x_0=x_0$ since $m_{-1}=0$. Expanding the update gives
\begin{align}
\widetilde x_{t+1}
&=x_t-\eta m_t-\frac{\eta\beta}{1-\beta}m_t\notag\\
&=x_t-\frac{\eta}{1-\beta}
 [\beta m_{t-1}+(1-\beta)g_t]
=\widetilde x_t-\eta g_t.
\label{eq:sgdm_virtual_update}
\end{align}
Hence, by using $L$-smoothness at $\widetilde x_t$, we obtain
\begin{equation}\label{eq:L_smooth_at_tilda_x}
f(\widetilde x_{t+1})
\le f(\widetilde x_t)
-\eta\langle\nabla f(\widetilde x_t),g_t\rangle
+\frac{L\eta^2}{2}\norm{g_t}^2.
\end{equation}
Taking conditional expectation $\E_t[\cdot]$ and using
$$
\E_t[g_t]=\nabla f(x_t),\qquad
\E_t\norm{g_t}^2=\norm{\nabla f(x_t)}^2+\E_t\norm{e_t}^2,
$$
along with the polarization identity and using $L$-smoothness, we have
\begin{align}
-\langle\nabla f(\widetilde x_t),\nabla f(x_t)\rangle
&\le-\frac12\norm{\nabla f(x_t)}^2
+\frac12\norm{\nabla f(\widetilde x_t)-\nabla f(x_t)}^2\notag\\
&\le-\frac12\norm{\nabla f(x_t)}^2
+\frac{L^2\eta^2\beta^2}{2(1-\beta)^2}\norm{m_{t-1}}^2.
\end{align}
Applying the previous steps to \eqref{eq:L_smooth_at_tilda_x}, we arrive at
\begin{align}\label{eq:sgdm_descent}
\E_t[f(\widetilde x_{t+1})]
&\le f(\widetilde x_t)
-\left(\frac\eta2-\frac{L\eta^2}{2}\right)\norm{\nabla f(x_t)}^2
+\frac{L^2\eta^3\beta^2}{2(1-\beta)^2}\norm{m_{t-1}}^2
+\frac{L\eta^2}{2}\E_t\norm{e_t}^2.
\end{align}

\textbf{Step 2: Telescoping to obtain the stationarity criterion.}
We first control the additional momentum term in \eqref{eq:sgdm_descent}. Conditional unbiasedness and convexity of the squared norm imply
\begin{align}
\E_t\norm{m_t}^2
&=\norm{\beta m_{t-1}+(1-\beta)\nabla f(x_t)}^2
 +(1-\beta)^2\E_t\norm{e_t}^2\notag\\
&\le\beta\norm{m_{t-1}}^2
 +(1-\beta)\norm{\nabla f(x_t)}^2
 +(1-\beta)^2\E_t\norm{e_t}^2.
\end{align}
Take expectations and sum over $t=0,\ldots,T-2$. The left side is
$\sum_{t=0}^{T-1}\E\norm{m_{t-1}}^2$, since $m_{-1}=0$;
the momentum sum on the right is at most this same sum multiplied by $\beta$.
Moving it to the left, dividing by $1-\beta$, and enlarging the remaining sums gives
\begin{equation}\label{eq:sgdm_momentum_energy}
\sum_{t=0}^{T-1}\E\norm{m_{t-1}}^2
\le\sum_{t=0}^{T-1}\E\norm{\nabla f(x_t)}^2
 +(1-\beta)\sum_{t=0}^{T-1}\E\norm{e_t}^2.
\end{equation}
For $T=1$, the same inequality holds because the left side is zero.
Next, we take total expectations in \eqref{eq:sgdm_descent}, sum over $t=0,\ldots,T-1$, and use
$f(\widetilde x_T)\ge f^{\inf}$ with $\widetilde x_0=x_0$. Substituting \eqref{eq:sgdm_momentum_energy} gives
\begin{align}\label{eq:sgdm_telescope}
\left(\frac\eta2-\frac{L\eta^2}{2}-A\right)
\sum_{t=0}^{T-1}\E\norm{\nabla f(x_t)}^2
\le\Delta+
\left(A(1-\beta)+\frac{L\eta^2}{2}\right)
\sum_{t=0}^{T-1}\E\norm{e_t}^2,
\end{align}
where we have denoted $A:=\frac{L^2\eta^3\beta^2}{2(1-\beta)^2}$ for convenience. Applying the restriction on the step size $\eta\le(1-\beta)^2/(2L)$, we get
\begin{align*}
\frac{L\eta^2}{2}&\le\frac\eta4,\qquad
A\le\frac{\eta\beta^2(1-\beta)^2}{8}\le\frac\eta8,\\
A(1-\beta)
&=\frac{L\eta^2}{2}\frac{L\eta\beta^2}{1-\beta}
\le\frac{L\eta^2}{2}\frac{\beta^2(1-\beta)}2
\le\frac{L\eta^2}{2}.
\end{align*}
Thus the coefficient on the left of \eqref{eq:sgdm_telescope} is at least $\eta/8$, whereas the coefficient of the error sum is at most $L\eta^2$. Dividing and using \eqref{eq:sgdm_cap} yields
\begin{equation}\label{eq:sgdm_gradient_sum}
\sum_{t=0}^{T-1}\E\norm{\nabla f(x_t)}^2
\le\frac{8\Delta}{\eta}
+\frac{8L\eta T\sigma^2}{\cB}.
\end{equation}
Finally, dividing by $T$ and substituting \eqref{eq:sgdm_convergence_parameters}
$$
\frac1T\sum_{t=0}^{T-1}\E\norm{\nabla f(x_t)}^2
\le\frac{8\Delta}{\eta T}+\frac{8L\eta\sigma^2}{\cB}
\le\frac{\eps^2}{2}+\frac{\eps^2}{2}=\eps^2,
$$
which proves \eqref{eq:sgdm_accuracy}.

\textbf{Step 3: SFO computation.}
For $1\le t\le T$, unrolling the momentum recursion and then summing the iterate updates gives
\begin{align}
m_j&=(1-\beta)\sum_{i=0}^j\beta^{j-i}g_i,\notag\\
x_t-x_0
&=-\eta(1-\beta)\sum_{j=0}^{t-1}\sum_{i=0}^j\beta^{j-i}g_i
=-\eta\sum_{i=0}^{t-1}(1-\beta^{t-i})g_i.
\label{eq:sgdm_displacement}
\end{align}
For fixed $t$, denote $w_{t,i}=1-\beta^{t-i}\in[0,1]$, which are deterministic constants. Hence, splitting $g_i=\nabla f(x_i)+e_i$ gives
\begin{equation}\label{eq:sgdm_split}
\E\norm{x_t-x_0}^2
\le2\eta^2\E\left\|\sum_{i=0}^{t-1}w_{t,i}\nabla f(x_i)\right\|^2
+2\eta^2\E\left\|\sum_{i=0}^{t-1}w_{t,i}e_i\right\|^2.
\end{equation}
Cauchy--Schwarz bounds the first term inside its expectation by
$$
\E\left\|\sum_{i=0}^{t-1}w_{t,i}\nabla f(x_i)\right\|^2
\le t\sum_{i=0}^{t-1}\E\norm{\nabla f(x_i)}^2
\le T\sum_{i=0}^{T-1}\E\norm{\nabla f(x_i)}^2.
$$
For $i<j$, $e_i$ is $\cF_j$-measurable and
$$
\E\langle e_i,e_j\rangle
=\E\bigl[\langle e_i,\E[e_j\mid\cF_j]\rangle\bigr]=0.
$$
The deterministic weights therefore preserve martingale orthogonality, and \eqref{eq:sgdm_cap} gives
$$
\E\left\|\sum_{i=0}^{t-1}w_{t,i}e_i\right\|^2
=\sum_{i=0}^{t-1}w_{t,i}^2\E\norm{e_i}^2
\le\frac{T\sigma^2}{\cB}.
$$
Substituting these bounds and \eqref{eq:sgdm_gradient_sum} into \eqref{eq:sgdm_split} yields
\begin{align}\label{eq:sgdm_radius}
\E\norm{x_t-x_0}^2
&\le2\eta^2T\left(\frac{8\Delta}{\eta}
+\frac{8L\eta T\sigma^2}{\cB}\right)
+\frac{2\eta^2T\sigma^2}{\cB}\notag\\
&=16\eta T\Delta
+\frac{16L\eta^3T^2\sigma^2}{\cB}
+\frac{2\eta^2T\sigma^2}{\cB}
=:D_T.
\end{align}
The same bound holds at $t=0$ because the displacement is zero.

\textbf{Step 4: $a$ power growth.}
For $0 \leq a \le 2$, the map $u\mapsto u^{a/2}$ is concave on $[0,\infty)$. Hence, Jensen's inequality and \eqref{eq:sgdm_radius} imply
\begin{equation}\label{eq:sgdm_power_growth_jensen}
\E\norm{x_t-x_0}^a
=\E\left[(\norm{x_t-x_0}^2)^{a/2}\right]
\le\left(\E\norm{x_t-x_0}^2\right)^{a/2}
\le D_T^{a/2}.
\end{equation}
Using $\max\{1,\lceil u\rceil\}\le1+u$ for $u\ge0$, summing the batch sizes, and taking total expectation gives
\begin{align}
\E\left[\sum_{t=0}^{T-1}b_t\right]
&\le\sum_{t=0}^{T-1}\left[
1+\frac{\cB G^2}{\sigma^2}
+\frac{\cB B^2}{\sigma^2}\E[\norm{x_t-x_0}^a]\right]\notag\\
&\le T\left(1+\frac{\cB G^2}{\sigma^2}
+\frac{\cB B^2}{\sigma^2}D_T^{a/2}\right).
\label{eq:sgdm_exp_of_bt}
\end{align}

\textbf{Step 5: Substituting the explicit constants.}
The restriction $\eps^2\le\min\{8\sigma^2(1-\beta^2),L\Delta\}$ that we set gives
$$
\frac{\cB\eps^2}{16L\sigma^2}
\le\frac{\cB(1-\beta^2)}{2L}
=\frac{(1-\beta)^2}{2L}.
$$
Hence $\eta \leq (1-\beta)^2/(2L)$ as
$$
\eta=\frac{\cB\eps^2}{16L\sigma^2},\qquad
\frac{16\Delta}{\eta\eps^2}
=\frac{256L\sigma^2\Delta}{\cB\eps^4}.
$$
Since $\eta\le1/(2L)$ and $\eps^2\le L\Delta$, the expression inside the ceiling for $T$ is at least $32$, and hence at least $4$. Applying $\lceil u\rceil\le(5/4)u$ for $u\ge4$ yields
\begin{equation}\label{eq:sgdm_Tbound}
T\le\frac{320L\sigma^2\Delta}{\cB\eps^4},\qquad
\eta T\le\frac{20\Delta}{\eps^2}.
\end{equation}
Next, we bound the three terms of $D_T$ separately
\begin{align}
16\eta T\Delta&\le\frac{320\Delta^2}{\eps^2},\\
\frac{16L\eta^3T^2\sigma^2}{\cB}
=\frac{16L\eta\sigma^2}{\cB}(\eta T)^2
&\le\eps^2\frac{400\Delta^2}{\eps^4}
=\frac{400\Delta^2}{\eps^2},\\
\frac{2\eta^2T\sigma^2}{\cB}
=\frac{2\eta\sigma^2}{\cB}(\eta T)
&\le\frac{\eps^2}{8L}\frac{20\Delta}{\eps^2}
=\frac{5\Delta}{2L}
\le\frac{5\Delta^2}{2\eps^2}.
\end{align}
The last inequality uses $\eps^2\le L\Delta$. Therefore,
\begin{equation}\label{eq:sgdm_sharpD}
D_T\le\frac{725\Delta^2}{\eps^2}.
\end{equation}
Combining \eqref{eq:sgdm_Tbound}, \eqref{eq:sgdm_sharpD}, and \eqref{eq:sgdm_exp_of_bt} yields
\begin{align*}
\E[\mathrm{SFO}]
&\le\frac{320L\sigma^2\Delta}{\cB\eps^4}
\left[1+\frac{\cB G^2}{\sigma^2}
+\frac{\cB B^2}{\sigma^2}
\left(\frac{725\Delta^2}{\eps^2}\right)^{a/2}\right]\\
&=320L\Delta\left[
\left(\frac{\sigma^2}{\cB}+G^2\right)\eps^{-4}
+725^{a/2}B^2\Delta^a\eps^{-(4+a)}\right].
\end{align*}
This proves \eqref{eq:sgdm_explicitcost_} and completes the proof.
\end{proof}

\textbf{Interpretation of \eqref{eq:sgdm_explicitcost_}.}
For fixed $\beta$ and $\sigma^2$, the dependence on the distance exponent is
\begin{center}
\begin{tabular}{@{}lll@{}}\toprule
Distance exponent & Variance upper bound & Expected SFO upper bound\\\midrule
$a=0$ & $B^2+G^2$ & $\Ocal(\eps^{-4})$\\
$0<a<2$ & $B^2\norm{x-x_0}^a+G^2$ & $\Ocal(\eps^{-(4+a)})$\\
$a=2$ & $B^2\norm{x-x_0}^2+G^2$ & $\Ocal(\eps^{-6})$\\\bottomrule
\end{tabular}
\end{center}
When $\beta=0$, we get $\cB=1$ and $m_t=g_t$, so the update and batch rule reduce to SGD. When $a=2$, we achieve the $\Ocal(\eps^{-6})$ rate from \cite{fazla2026lower}. When $B=0$, the batch size is constant and the bound reduces to the classical $\Ocal(\eps^{-4})$ rate.

\textbf{Momentum as a batch size and iteration tradeoff.}
The momentum coefficient $\beta$ acts as a hyperparameter controlling the allocation of oracle calls between the batch size and the number of iterations $T$. Since $\cB=(1-\beta)/(1+\beta)$ decreases as $\beta$ increases, the batch rule in \eqref{eq:sgdm_convergence_parameters} prescribes a smaller batch at the same query point. This is observed in the proof, as the momentum filter reduces the noise second moment by the factor $\cB$, allowing the minibatch estimator to have variance at most $\sigma^2/\cB$ while its filtered noise component has second moment at most $\sigma^2$.

Under the theorem's accuracy restriction, the learning rate and iteration budget satisfy
$$
\eta=\frac{\cB\eps^2}{16L\sigma^2},\qquad
T=\left\lceil\frac{256L\sigma^2\Delta}{\cB\eps^4}\right\rceil.
$$
Thus a larger $\beta$ permits smaller prescribed batches but requires a smaller learning rate and more iterations. Relative to $\beta=0$ within this SGDM parameterization, the unrounded batch target is multiplied by $\cB$, whereas the expression inside the ceiling for $T$ is multiplied by $1/\cB$. The batch comparison is at a common query point, since different momentum choices generally produce different trajectories. In \eqref{eq:sgdm_explicitcost_}, these factors cancel in the $B^2$ and $G^2$ terms, while the minimum cost of one sample per iteration retains the factor $1/\cB$. Hence $\beta$ controls a batch size/iteration tradeoff while preserving the $\eps$ exponent for fixed $\beta$ and the bound does not provide a reduction in total SFO cost as momentum increases. Note that the accuracy restriction must also hold for the chosen $\beta$, and the expected SFO statement is restricted to $0\le a\le2$.

\section{Additional Experiments}\label{app:add_exps}

\subsection{LLM Experiments}
We present the downstream accuracy scores on HellaSwag and CommonsenseQA for the three OLMo2 models (100M, 600M, 1B) in Table~\ref{tab:olmo2_downstream}. Each model is trained with 20 tokens per parameter, corresponding to 2B, 12B, and 20B training tokens, respectively. We observe that BG-$a$ attains the highest mean accuracy on both benchmarks, with the clearest gains over small batch training at the larger model sizes, suggesting that the improvement in validation loss carries over to downstream tasks.

\begin{table}[!t]
\centering
\caption{Downstream performance of OLMo2 models pretrained
on C4. Entries are accuracy (\%),
reported as mean$\pm$std across $3$ seeds.}
\label{tab:olmo2_downstream}
\begin{tabular}{llrr}
\toprule
Model & Batch Size & HellaSwag & CommonsenseQA \\
\midrule
\multirow{3}{*}{OLMo2-100M}
 & Small ($b=32$) & $28.98 \pm 0.08$ & $29.03 \pm 0.06$ \\
 & Large ($b=512$) & $27.97 \pm 0.25$ & $27.27 \pm 0.01$ \\
 & BG-$a$            & $29.15 \pm 0.30$ & $29.07 \pm 0.46$ \\
\midrule
\multirow{3}{*}{OLMo2-600M}
 & Small ($b=32$) & $ 40.43 \pm 0.11 $ & $ 32.43 \pm 0.04$ \\
 & Large ($b=512$) & $ 43.05 \pm 0.15$ & $ 32.92 \pm 0.11$ \\
 & BG-$a$            & $ 43.11 \pm 0.32$ & $ 33.33 \pm 0.34$ \\
\midrule
\multirow{3}{*}{OLMo2-1B}
 & Small ($b=64$) & $ 50.90 \pm 0.12 $ & $ 33.66 \pm 0.09 $ \\
 & Large ($b=2048$) & $ 51.12 \pm 0.47 $ & $ 34.89 \pm 0.22 $ \\
 & BG-$a$            & $ 51.86 \pm 0.26 $ & $ 36.53 \pm 0.41 $ \\
\bottomrule
\end{tabular}
\end{table}

\subsection{Vision Experiments}\label{app:vis_exps}

Here, we provide our experiments on image classification with the ResNet50 architecture on CIFAR-10, CIFAR-100 and ImageNet-1K. We first repeat the noise growth study of Section~\ref{sec:variance_ablation} for these models, and then evaluate our adaptive batch size scheduler. Unlike LLM pretraining, these models are trained for multiple epochs over a fixed dataset, which allows us to examine how the optimization benefits of adaptive batching translate to train and test performances in the offline setting.

\subsubsection{Gradient Noise Growth in Image Classification}\label{app:vis_noise}

We repeat the noise growth study of Section~\ref{sec:variance_ablation} for ResNet50 on CIFAR-10, CIFAR-100, and ImageNet-1K. Unlike the language models, we set $D=I$ for the vision models, so the perturbation direction $u$ is uniformly distributed on the unit sphere. As in the language model study, we consider $50$ logarithmically spaced radii and sample $10$ independent directions at each radius. For CIFAR-10 and CIFAR-100, the reference batch contains $5000$ images, and the estimate uses $M=16$ stochastic batches, each containing $b=128$ images. For ImageNet-1K, we use $1024$ reference images and $M=32$ stochastic batches of $b=16$ images.

\begin{figure}[!t]
    \centering
    \includegraphics[width=0.9\textwidth]{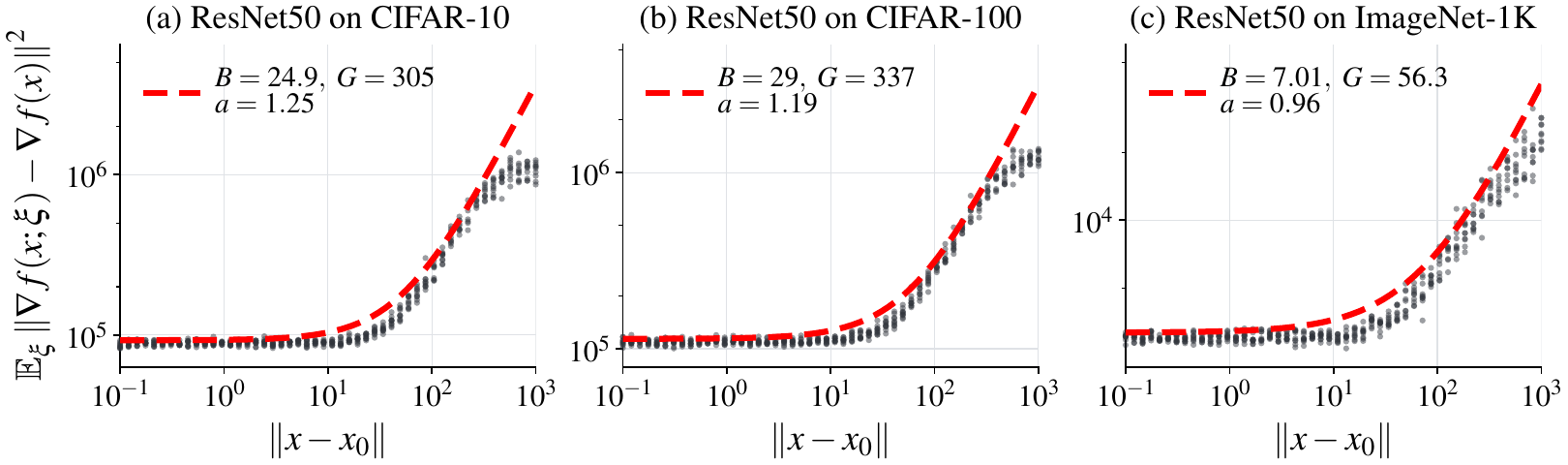}
    \caption{
    Empirical stochastic-gradient noise of ResNet50 as a function of displacement from initialization. For each radius $r$, we evaluate $x=x_0+ru$ along ten independently sampled, isotropic Gaussian directions and plot the corresponding minibatch variance estimates. The dashed red curve in each panel is the fitted BG-$a$ model, ($B^2\norm{x-x_0}^a+G^2$). The fitted parameters are: \textbf{(a)} CIFAR-10: $B=24.9$, $G=305$, $a=1.25$; \textbf{(b)} CIFAR-100: $B=29$, $G=337$, $a=1.19$; and \textbf{(c)} ImageNet-1K: $B=7.01$, $G=56.3$, $a=0.96$.
    }
    \label{fig:resnet50_noise}
\end{figure}

Figure~\ref{fig:resnet50_noise} shows the variance measurements together with the fitted BG-$a$ curves. The fitted exponents are approximately $a=1.25$ for CIFAR-10, $a=1.19$ for CIFAR-100, and $a=0.96$ for ImageNet-1K. Similar to the language models, the variance grows subquadratically with the distance from initialization, and the BG-$a$ model captures both the initial noise floor ($G^2$ term) and the distance-dependent growth (from $B^2$ term).

\subsubsection{Experimental Results}\label{app:vis_training}

\begin{figure}[t]
    \centering
    \includegraphics[width=\textwidth]{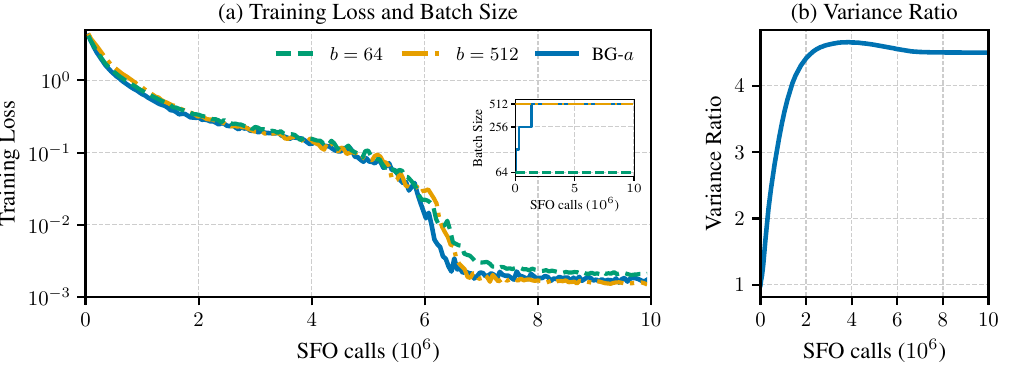}
    \caption{
        Results for ResNet50 trained on CIFAR-100.
        (a) Training loss for fixed batches $b=64$, $b=512$, and
        BG-$a$, with batch size across iterations shown in the inset.
        (b) Modeled variance ratio for BG-$a$, with parameters $a=1.19, B=29, G=337$
    }
    \label{fig:cifar100}
\end{figure}

\begin{figure}[t]
    \centering
    \includegraphics[width=\textwidth]{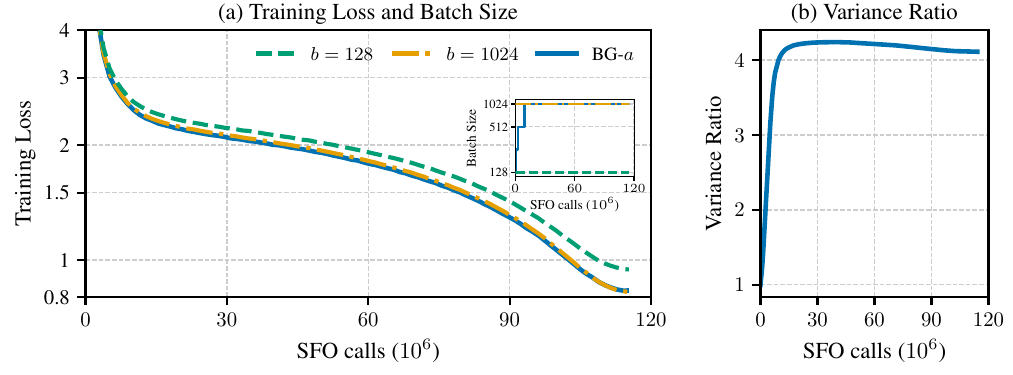}
    \caption{
        Results for ResNet50 trained on ImageNet-1K.
        (a) Training loss for fixed batches $b=128$, $b=1024$, and
        BG-$a$, with batch size across iterations shown in the inset.
        (b) Modeled variance ratio for BG-$a$, with parameters $a=0.96, B=7.01, G=56.3$
    }
    \label{fig:imagenet}
\end{figure}

We compare our adaptive batch size scheduler, denoted as BG-$a$, against a small and a large fixed batch baseline under the same number of training epochs. All runs use SGD with momentum, and the remaining training details are provided in Appendix~\ref{app:exp_details}. Figures~\ref{fig:cifar10}, \ref{fig:cifar100} and \ref{fig:imagenet} show the training loss curves, with the batch sizes selected by BG-$a$ shown in the insets, together with the modeled variance ratio. We report the final training and test losses in Table~\ref{tab:vision_results}.

Across all three datasets, BG-$a$ achieves a lower training loss than small batch training, while using only $14$--$16\%$ of its iterations. Compared to large batch training, BG-$a$ attains a lower training loss on CIFAR-10 and a slightly higher training loss on CIFAR-100 and ImageNet-1K, with a similar number of iterations. The improvement over small batch training is in line with our theoretical analysis: BG-$a$ controls the variance growth as the iterates move away from initialization, whereas small batch training does not and attains the highest training loss in all three settings.

However, unlike in LLM pretraining, where each batch consists of previously unseen tokens and the validation loss directly reflects the optimized objective, a lower training loss does not directly translate into a lower test loss in image classification. On CIFAR-10 and CIFAR-100, small batch training attains the lowest test loss despite having the highest training loss, while BG-$a$ improves upon large batch training in test loss on both datasets. This contrast is consistent with the distinction between offline and online learning discussed in Appendix~\ref{app:related_work}. In multi-epoch training on small datasets such as CIFAR, the models nearly interpolate the training data, with training losses of order $10^{-4}$ and $10^{-3}$, and the implicit regularization effect of small batches becomes the dominant factor for generalization \citep{keskar2016large}. In contrast, \citet{vyas2023beyond} show that this implicit bias advantage disappears in the online setting, where each sample is seen at most once and the benefit of small batches becomes purely computational. LLM pretraining falls into this online regime, where improving the training objective directly improves the validation loss, as observed in Section~\ref{sec:experiments}. ImageNet-1K lies between these two cases: its larger dataset keeps the gap between training and test losses small, and the ordering of test losses closely follows the ordering of training losses. We also note that, analysis of stochastic nonconvex optimization often assumes that each iteration draws fresh and independent samples, which holds in online learning (e.g., LLM pretraining) but not in multi-epoch training, where every sample is revisited after the first epoch, i.e., after only $0.5\%$ of training on CIFAR and $1.1\%$ on ImageNet-1K. The stochastic gradients are then no longer independent draws from the data distribution, and the noise model estimated at initialization need not describe the noise encountered later in training.

\begin{figure}[t]
    \centering
    \includegraphics[width=\textwidth]{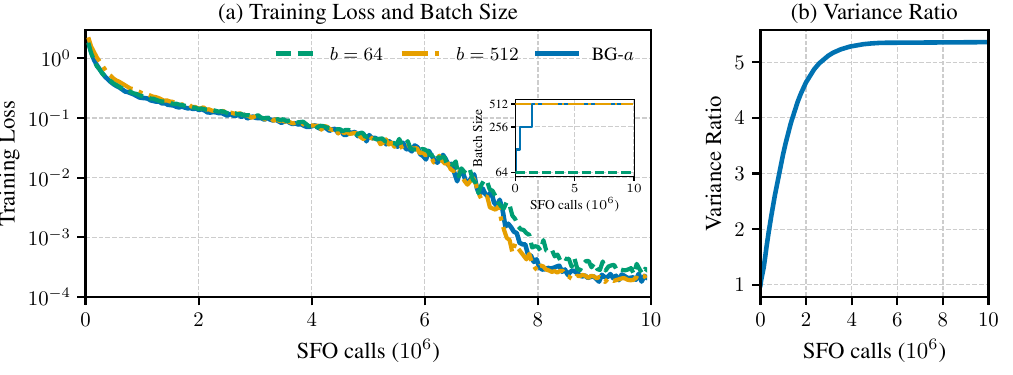}
    \caption{
        Results for ResNet50 trained on CIFAR-10.
        (a) Training loss for fixed batches $b=64$, $b=512$, and
        BG-$a$, with batch size across iterations shown in the inset.
        (b) Modeled variance ratio for BG-$a$, with parameters $a=1.25, B=24.9, G=305$
    }
    \label{fig:cifar10}
\end{figure}

\begin{table}[!t]
\centering
\small
\setlength{\tabcolsep}{3pt}
\caption{Training and test performance of the ResNet50 model trained
on CIFAR-10, CIFAR-100 and ImageNet-1K. Entries are loss,
reported as mean$\pm$std across $3$ seeds. Iterations are averaged 
across seeds and given as total iterations and percentage relative 
to the small-batch baseline within each dataset.}
\label{tab:vision_results}
\begin{tabular}{@{}llrrrr@{}}
\toprule
Dataset & Batch Size & Train Loss & Test Loss
& \shortstack{Iterations\\($T$)}
& \shortstack{Iterations\\(\%)} \\
\midrule
\multirow{3}{*}{CIFAR-10}
& Small ($b=64$)
& $0.000262 \pm 0.000020$ & $0.2253 \pm 0.0086$
& 156,200 & 100.00 \\
& Large ($b=512$)
& $0.000204 \pm 0.000021$ & $0.2508 \pm 0.0028$
& 19,400 & 12.42 \\
& BG-$a$
& $0.000200 \pm 0.000009$ & $0.2412 \pm 0.0028$
& 23,800 & 15.24 \\
\midrule
\multirow{3}{*}{CIFAR-100}
& Small ($b=64$)
& $0.002215 \pm 0.000128$ & $0.9522 \pm 0.0052$
& 156,200 & 100.00 \\
& Large ($b=512$)
& $0.001496 \pm 0.000009$ & $0.9951 \pm 0.0244$
& 19,400 & 12.42 \\
& BG-$a$
& $0.001746 \pm 0.000193$ & $0.9609 \pm 0.0160$
& 23,440 & 15.00 \\
\midrule
\multirow{3}{*}{ImageNet-1K}
& Small ($b=128$)
& $0.9445 \pm 0.0011$ & $0.9260 \pm 0.0032$
& 900,810 & 100.00 \\
& Large ($b=1024$)
& $0.8227 \pm 0.0009$ & $0.9141 \pm 0.0017$
& 112,590 & 12.50 \\
& BG-$a$
& $0.8316 \pm 0.0008$ & $0.9255 \pm 0.0067$
& 127,100 & 14.11 \\
\bottomrule
\end{tabular}
\end{table}

\section{Experimental Details}\label{app:exp_details}

\paragraph{Models and computational resources.}
For all vision experiments, we use the ResNet50 architecture \citep{he2016deep}. For ImageNet-1K \citep{deng2009imagenet}, we use the standard PyTorch implementation without architectural modifications. For CIFAR-10 and CIFAR-100 \citep{alex2009learning}, we replace the initial $7\times7$ convolution with a $3\times3$ convolution to accommodate for their $32\times32$ image sizes. For language model pretraining, we use the OLMo2 family and the OLMo-core repository \citep{olmo20242olmo2furious} at commit \texttt{92870a3}. Model architectures are reported in Table~\ref{tab:olmo2_architectures}. Main vision experiments use four NVIDIA A100 GPUs, while main language modeling experiments use four NVIDIA H100 GPUs. We use distributed data parallelism (DDP) for all vision models and for OLMo2-100M and OLMo2-600M, and fully sharded data parallelism (FSDP) for OLMo2-1B.

\paragraph{Vision training.}
We train ResNet50 for $200$ epochs on CIFAR-10 and CIFAR-100 and for $90$ epochs on ImageNet-1K, using standard data augmentation with random resized cropping. All runs use SGD with momentum $0.9$ (SGDM) and weight decay $10^{-4}$. We select the base learning rate through a grid search at the small fixed batch size and scale it linearly for larger fixed batch sizes. The learning rate follows a linear warmup over the first five epochs, followed by cosine decay to zero. Table~\ref{tab:repro_vision} reports the configurations and mean training times.

\begin{table}[t]
\centering
\caption{
Vision training configurations and mean wall-clock times,
using four NVIDIA A100 GPUs. For BG-$a$, the batch size and base learning rate refer to the initial batch size and its associated base learning rate. ($*$) means that the given values are the initial starting values.
}
\label{tab:repro_vision}
\small
\setlength{\tabcolsep}{5pt}
\begin{tabular}{llrrrr}
\toprule
Dataset & Method & Epochs & Batch size & Base LR & Time (min) \\
\midrule
CIFAR-10
& Small batch & 200 & 64  & 0.05   & 338 \\
& Large batch & 200 & 512 & 0.4    & 206 \\
& BG-$a$      & 200 & 64$^{*}$  & 0.05$^{*}$   & 240 \\
\midrule
CIFAR-100
& Small batch & 200 & 64  & 0.05   & 337 \\
& Large batch & 200 & 512 & 0.4    & 206 \\
& BG-$a$      & 200 & 64$^{*}$  & 0.05$^{*}$   & 240 \\
\midrule
ImageNet-1K
& Small batch & 90 & 128  & 0.05 & 761 \\
& Large batch & 90 & 1024 & 0.4  & 489 \\
& BG-$a$      & 90 & 128$^{*}$  & 0.05$^{*}$ & 542 \\
\bottomrule
\end{tabular}
\end{table}

\paragraph{Language model pretraining.}
We pretrain OLMo2-100M, OLMo2-600M, and OLMo2-1B on C4 using
Chinchilla optimal \citep{hoffmann2022training} training budgets of approximately $20$ tokens per parameter, corresponding to $2$B, $12$B, and $20$B tokens, respectively. All runs use Adam with decoupled weight decay (AdamW), with
$\beta_1=0.9$, $\beta_2=0.95$, and weight decay $0.1$, following the
repository defaults. We set the sequence length to $2048$. We select the base learning rate through a grid search at the small fixed batch size and apply square root scaling for larger fixed batch sizes. The learning rate increases linearly to its base value over the first $5\%$ of training and subsequently follows a cosine schedule to $10\%$ of that value. Table~\ref{tab:repro_lm} reports the learning rates used, including the initial base learning rates for BG-$a$.

\begin{table}[t]
\centering
\caption{
Language model pretraining configurations on C4 and mean wall-clock
times, using four NVIDIA H100 GPUs. For BG-$a$, the batch size and base learning rate refer to the initial batch size and its associated base learning rate. ($*$) means that the given values are the initial starting values.
}
\label{tab:repro_lm}
\small
\setlength{\tabcolsep}{4pt}
\begin{tabular}{llrrrl}
\toprule
Model & Method & Tokens & Batch size & Base LR & Time \\
\midrule
OLMo2-100M
& Small batch & 2B & 32  & 0.004 & 28m 37s \\
& Large batch & 2B & 512 & 0.016 & 16m 13s \\
& BG-$a$      & 2B & 32$^{*}$  & 0.004$^{*}$ & 17m 20s \\
\midrule
OLMo2-600M
& Small batch & 12B & 32   & 0.001 & 12h 58m \\
& Large batch & 12B & 512 & 0.004 & 9h 03m \\
& BG-$a$      & 12B & 32$^{*}$   & 0.001$^{*}$  & 9h 17m \\
\midrule
OLMo2-1B
& Small batch & 20B & 64   & 0.001 & 36h 12m \\
& Large batch & 20B & 2048 & 0.0057  & 27h 35m \\
& BG-$a$      & 20B & 64$^{*}$   & 0.001$^{*}$ & 29h 05m \\
\bottomrule
\end{tabular}
\end{table}

\paragraph{Selection of the initial batch size and learning rate.}
For each setting, we tune the base learning rate through a grid search over $\{0.005, 0.01, 0.025, 0.05, 0.1\}$ for vision models and $\{0.0001, 0.0005, 0.001, 0.0025, 0.004, 0.005\}$ for language models, for each candidate small batch size in $\{32, 64, 128\}$ for vision models and $\{16, 32, 64\}$ for language models. We select the smallest batch size that trains stably and attains a competitive final loss. The selected batch size and its tuned learning rate are used both by the small batch baseline and as the initial batch size $b_0$ and base learning rate of BG-$a$. The large batch baseline uses the same base learning rate, scaled linearly for SGDM and with the square root rule for AdamW, without additional tuning. The resulting values are reported in Tables~\ref{tab:repro_vision} and~\ref{tab:repro_lm}.

\paragraph{Adaptive batching and learning rate scaling.}
BG-$a$ uses the noise parameters estimated by the procedure in
Section~\ref{sec:variance_ablation}, with fitted values reported in
Figure~\ref{fig:olmo2_row}. Following
Algorithm~\ref{alg:adaptive_batching_optimizer}, the scheduler selects
$b_t=k_tb_0$, where $k_t$ is the power-of-two multiplier defined in \eqref{eq:adaptive_batch_size_function}.
The warmup and cosine schedule defines $\eta_t$ at the initial batch
size $b_0$. At each iteration, we apply the batch dependent adjustment
in \eqref{eq:batch_lr_scaling}, hence, the effective learning rate is
$k_t\eta_t$ for the vision experiments that utilize SGDM and $\sqrt{k_t}\eta_t$ for language model pretraining which utilize AdamW. Thus, the learning rate schedule and the batch dependent scaling are applied jointly throughout training.

\paragraph{BG-$a$ noise growth ablations.}
We perform the ablation studies using the same repositories as their corresponding training experiments. As described in Section~\ref{sec:variance_ablation}, these studies evaluate stochastic gradient variance at randomly perturbed parameter states without performing training. Each study uses $50$ logarithmically spaced radii and $10$ independently sampled directions per radius. For language models, we multiply the Gaussian coordinates corresponding to the LM head by two, leave the remaining coordinates unchanged, and then normalize the full direction to unit Euclidean norm. For vision models, we normalize the unscaled Gaussian direction. In both cases, the perturbation radius equals the total Euclidean displacement from initialization. OLMo2-600M and OLMo2-1B ablations use four NVIDIA H100 GPUs, and all other ablations use one NVIDIA H100 GPU. Table~\ref{tab:repro_ablations} reports the wall-clock cost of each study.

\begin{table}[!t]
\centering
\caption{
Computational resources and wall-clock times for the BG-$a$ noise growth ablation studies.
}
\label{tab:repro_ablations}
\small
\begin{tabular}{llrl}
\toprule
Model & Dataset & H100 GPUs & Time \\
\midrule
ResNet50  & CIFAR-10    & 1 & 2m 44s \\
ResNet50  & CIFAR-100   & 1 & 2m 38s \\
ResNet50  & ImageNet-1K & 1 & 11m 49s \\
OLMo2-100M & C4          & 1 & 50m 08s \\
OLMo2-600M & C4          & 4 & 2h 34m 28s \\
OLMo2-1B   & C4          & 4 & 2h 56m 39s \\
\bottomrule
\end{tabular}
\end{table}

\begin{table}[!t]
\centering
\caption{
Architectural details of the OLMo2 models used in our experiments. 
All models use a decoder-only Transformer architecture with SwiGLU feed-forward layers, RMSNorm, QK-normalized attention, and rotary positional embeddings.
}
\label{tab:olmo2_architectures}
\begin{tabular}{lccc}
\toprule
 & OLMo2-100M & OLMo2-600M & OLMo2-1B \\
\midrule
Parameters      & 100M & 600M & 1B \\
Layers          & 12   & 16   & 16 \\
Hidden dimension     & 512  & 1344 & 2048 \\
Attention heads & 8    & 16   & 16 \\
Head dimension  & 64   & 84   & 128 \\
FFN hidden dimension & 2048 & 5376 & 8192 \\
\bottomrule
\end{tabular}
\end{table}

\section{Discussion}\label{sec:discussion}
Our analysis concerns the optimization of the training objective, namely finding approximate stationary points, and our experiments support this view in both language modeling and image classification. Whether an optimization improvement carries over to the test performance depends on the training regime. In LLM pretraining, each token is seen at most once, so the training and validation losses estimate the same population loss, and the implicit bias advantage of small batches disappears \citep{vyas2023beyond}. Accordingly, adaptive batching attains a lower validation loss than both small and large batch training. In multi-epoch image classification, adaptive batching consistently achieves a lower training loss than small batch training under matched training budgets, but the models revisit the same samples many times, and small batches may attain a lower test loss despite their higher training loss, consistent with their implicit regularization effect \citep{keskar2016large} (Appendix~\ref{app:vis_exps}). Hence, our results suggest that adaptive batching is particularly well suited for the online regime, which matches the setting of modern LLM pretraining.

Our measurements in Section~\ref{sec:variance_ablation} show that the growth exponent $a$ increases with model size, from $a=0.98$ for OLMo2-100M to $a=1.49$ for OLMo2-1B. Under our analysis, a faster variance growth increases the oracle complexity and makes controlling the variance growth more important, which is consistent with small batch training falling behind both large batch training and BG-$a$ for OLMo2-600M and OLMo2-1B. Studying whether this trend continues at even larger scales, and how it relates to the critical batch size \citep{zhang2025does,merrill2026critical}, is an interesting direction for future work. Furthermore, \cite{zhang2025does} empirically show that CBS grows primarily with the training data size. Since we train with $20$ tokens per parameter, a larger model requires larger data. Our analysis and empirical results also suggest that larger models eventually require larger batch sizes during training, which is in accordance with the results of \cite{zhang2025does}.

Our scheduler follows the batch size rule of our analysis, while also incorporating common practices of large scale training. In particular, the scheduler scales the learning rate with the batch size, whereas our theorems use a fixed step size, and our language model experiments use AdamW, whereas our analysis covers SGD and SGDM. Extending the BG-$a$ analysis to adaptive methods and to include learning rate schedulers is a natural next step, and would further clarify the interaction between the batch size, the learning rate, and distance-dependent noise. Similarly, we select the initial batch size empirically, since the target noise level $\sigma^2$ is not known a priori, which is also a common practice for batch size schedulers based on GNS \citep{mccandlish2018empirical, naganuma2026adaptive}.

\textbf{Limitations.} Our experiments cover OLMo2 models of up to 1B parameters trained on C4 with Chinchilla optimal budgets. Evaluating the scheduler at larger scales and on other pretraining corpora would further clarify its practical benefits. In addition, estimating the noise parameters requires a one time measurement before training, whose cost we report in Table~\ref{tab:repro_ablations}. Since this measurement does not involve training and depends only on the model and dataset, it can be reused across training runs. Finally, our upper bounds are worst-case guarantees that hold for $a\le2$, which covers all the exponents we observe in practice, while extending them to $a>2$ requires controlling higher moments of the displacement, as discussed in Appendix~\ref{sec:convergence_of_sgd}.

\end{document}